\documentclass[oneside,english]{article}

\usepackage[utf8]{inputenc}

\usepackage{hyperref}
\usepackage{url}

\usepackage{algorithm,algpseudocode}

\usepackage{mathtools}
\usepackage{amsmath,amssymb,amsthm}
\usepackage{thmtools}

\usepackage{graphicx}

\usepackage{todonotes}

\usepackage[charter]{mathdesign}
\usepackage[mathcal]{eucal}
\usepackage{bbm}

\usepackage[shortlabels]{enumitem}

\usepackage[margin=1.2 in]{geometry}
\usepackage{pdflscape}

\usepackage{multirow}
\usepackage{array}
\usepackage[numbers]{natbib}
\usepackage{authblk}

\usepackage{titlesec}
\titleformat{\section}
	{\normalfont\Large\bfseries\filcenter}{\thesection.}{1 ex}{}
\titleformat{\subsection}%[runin]
	{\normalfont\normalsize\bfseries}{\thesubsection.}{1 ex}{}
\titleformat{\subsubsection}[runin]
	{\normalfont\normalsize\bfseries\filcenter}{\thesubsubsection.}{1 ex}{}

\newcommand{\argmax}{\operatornamewithlimits{argmax}}

\newcommand{\Tr}{\operatorname{Tr}}

\newcommand{\Corr}{\operatorname{Corr}}

\newcommand{\logit}{\operatorname{logit}}
\newcommand{\odds}{\operatorname{odds}}

\newcommand{\Distr}{\operatorname{Distr}}
\newcommand{\Bern}{\operatorname{Bernoulli}}

\newcommand{\calD}{\mathcal{D}}

\newcommand{\calI}{\mathcal{I}}

\newcommand{\calT}{\mathcal{T}}

\newcommand{\calX}{\mathcal{X}}
\newcommand{\calY}{\mathcal{Y}}

\newcommand{\ZZ}{\ensuremath{\mathbb{Z}}}
\newcommand{\RR}{\ensuremath{\mathbb{R}}}

\newcommand{\NN}{\ensuremath{\mathbb{N}}}

\newcommand{\EE}{\ensuremath{\mathbb{E}}}

\newcommand{\OOne}{\ensuremath{\mathbbm{1}}}

\newcommand{\balpha}{\boldsymbol{\alpha}}
\newcommand{\bbeta}{\boldsymbol{\beta}}
\newcommand{\bgamma}{\boldsymbol{\gamma}}
\newcommand{\blambda}{\boldsymbol{\lambda}}

\newcommand{\bL}{\boldsymbol{L}}
\newcommand{\bN}{\boldsymbol{N}}
\newcommand{\bP}{\boldsymbol{P}}
\newcommand{\bW}{\boldsymbol{W}}
\newcommand{\bX}{\boldsymbol{X}}
\newcommand{\bY}{\boldsymbol{Y}}

\makeatletter
\providecommand*{\diff}%
        {\@ifnextchar^{\DIfF}{\DIfF^{}}}
\def\DIfF^#1{%
        \mathop{\mathrm{\mathstrut d}}%
                \nolimits^{#1}\gobblespace
}
\def\gobblespace{%
        \futurelet\diffarg\opspace}
\def\opspace{%
        \let\DiffSpace\!%
        \ifx\diffarg(%
                \let\DiffSpace\relax
        \else
                \ifx\diffarg\[%
                        \let\DiffSpace\relax
                \else
                        \ifx\diffarg\{%
                                \let\DiffSpace\relax
                        \fi\fi\fi\DiffSpace}
\makeatother

\makeatletter

\providecommand{\tabularnewline}{\\}
\floatstyle{ruled}
\newfloat{algorithm}{tbp}{loa}
\providecommand{\algorithmname}{Algorithm}
\floatname{algorithm}{\protect\algorithmname}

\makeatother

\usepackage{babel}

\begin{document}

\title{Modelling Competitive Sports:\\
Bradley-Terry-\'{E}l\H{o} Models\\ for Supervised and On-Line Learning\\ of Paired Competition Outcomes}

\author[1]{
Franz J.~Kir\'{a}ly
\thanks{\url{f.kiraly@ucl.ac.uk}}
}

\author[12]{Zhaozhi Qian
\thanks{\url{zhaozhi.qian.15@ucl.ac.uk}}
}

\affil[1]{
Department of Statistical Science,
University College London,\newline
Gower Street,
London WC1E 6BT, United Kingdom
}

\affil[2]{King Digital Entertainment plc,
Ampersand Building,\newline
178 Wardour Street,
London W1F 8FY, United Kingdom

}

\thispagestyle{empty}
\maketitle

%\newpage

\begin{abstract}
Prediction and modelling of competitive sports outcomes has received much recent attention,
especially from the Bayesian statistics and machine learning communities. In the real world setting of outcome prediction,
the seminal \'{E}l\H{o} update still remains, after more than 50 years, a valuable baseline
which is difficult to improve upon, though in its original form it is a heuristic and not a proper statistical ``model''.
Mathematically, the \'{E}l\H{o} rating system is very closely related to the Bradley-Terry models, which are usually
used in an explanatory fashion rather than in a predictive supervised or on-line learning setting.

Exploiting this close link between these two model classes and some newly observed similarities,
we propose a new supervised learning framework with close similarities to logistic regression, low-rank matrix completion and neural networks.
Building on it, we formulate a class of structured log-odds models, unifying the desirable properties found in the
above: supervised probabilistic prediction of scores and wins/draws/losses, batch/epoch and on-line learning, as well as the possibility to incorporate features in the prediction, without having to sacrifice simplicity, parsimony of the Bradley-Terry models, or computational efficiency of \'{E}l\H{o}'s original approach.

We validate the structured log-odds modelling approach in synthetic experiments and English Premier League outcomes,
where the added expressivity yields the best predictions reported in the state-of-art, close to the quality of contemporary betting odds.

\newpage
\end{abstract}
\tableofcontents{}

\newpage{}

\section{Introduction\label{sub:intro_one}}

%We give a short introduction to competitive team sports modelling, a literature overview, and a summary of our contributions.

\subsection{Modelling and predicting competitive sports\label{sub:intro_one}}

Competitive sports refers to any sport that involves two teams or individuals
competing against each other to achieve higher scores. Competitive
team sports includes some of the most popular and most watched games
such as football, basketball and rugby. Such sports are played both
in domestic professional leagues such as the National Basketball Association,
and international competitions such as the FIFA World Cup. For football
alone, there are over one hundred fully professional leagues in 71
countries globally. It is estimated that the Premier League, the top
football league in the United Kingdom, attracted a (cumulative) television
audience of 4.7 billion viewers in the last season~\citep{PremierLeagueAudience}.

The outcome of a match is determined by a large number of factors.
Just to name a few, they might involve the competitive strength of
each individual player in both teams, the smoothness of collaboration
between players, and the team's strategy of playing. Moreover, the
composition of any team changes over the years, for example because players leave
or join the team. The team composition may also change within the
tournament season or even during a match because of injuries or penalties.

Understanding these factors is, by the prediction-validation nature of the scientific method,
closely linked to predicting the outcome of a pairing. By Occam's razor, the
factors which empirically help in prediction are exactly those that one may hypothesize to
be relevant for the outcome.

Since keeping track of all relevant factors is unrealistic,
of course one cannot expect a certain prediction of a competitive sports outcome.
Moreover, it is also unreasonable to believe that all factors can be measured or controlled,
hence it is reasonable to assume that unpredictable, or non-deterministic statistical ``noise''
is involved in the process of generating the outcome (or subsume the unknowns as such noise).
A good prediction will, hence, not exactly predict the outcome, but will anticipate the ``correct'' odds more precisely.
The extent to which the outcomes are predictable may hence be considered as a surrogate quantifier of how much
the outcome of a match is influenced by ``skill'' (as surrogated by determinism/prediction), or by
``chance''\footnote{We expressly avoid use of the word ``luck'' as in vernacular use it often means ``chance'',
jointly with the belief that it may be influenced by esoterical, magical or otherwise metaphysical means.
While in the suggested surrogate use, it may well be that the ``chance'' component of a model subsumes
possible points of influence which simply are not measured or observed in the data, an
extremely strong corpus of scientific evidence implies that these will not be metaphysical, only unknown
- two qualifiers which are obviously not the same, despite strong human tendencies to believe the contrary.}
(as surrogated by the noise/unknown factors).

Phenomena which can not be specified deterministically
are in fact very common in nature. Statistics and probability theory provide ways
to make inference under randomness. Therefore, modelling and predicting
the results of competitive team sports naturally falls into the area
of statistics and machine learning. Moreover, any interpretable predictive model
yields a possible explanation of what constitutes factors influencing the outcome.

\subsection{History of competitive sports modelling}

Research of modeling competitive sports has a long history. In its early
days, research was often closely related to sports betting or player/team ranking~\citep{griffith1949odds,isaacs1953optimal}.
The two most influential approaches are due to~\citet{bradley1952rank} and~\citet{elo1978rating}.
The Bradley-Terry and  \'{E}l\H{o} models allow estimation of player rating;
the \'{E}l\H{o} system additionally contains algorithmic heuristics to easily update a player's rank,
which have been in use for official chess rankings since the 1960s.
The \'{E}l\H{o} system is also designed to predict the odds of a player winning or losing to the opponent.
In contemporary practice, Bradley-Terry and  \'{E}l\H{o} type models are broadly used in modelling of sports outcomes and ranking of players,
and it has been noted that they are very close mathematically.

In more recent days, relatively diverse modelling approaches originating from the Bayesian statistical framework
~\citep{maher1982modelling,dixon1997modelling,glickman1998state},
and also some inspired by machine learning principles~\citep{liu2010beating,hucaljuk2011predicting,odachowski2012using} have been applied for modelling competitive sports.
These models are more expressive and remove some of the Bradley-Terry and \'{E}l\H{o} models' limitations, though usually at the price of
interpretability, computational efficiency, or both.

A more extensive literature overview on existing approaches will be given later in Section~\ref{sub:A-brief-summary}, as literature
spans multiple communities and, in our opinion, a prior exposition of the technical setting
and simultaneous straightening of thoughts benefits the understanding and allows us
to give proper credit and context for the widely different ideas employed in competitive sports modelling.

\subsection{Aim of competitive sports modelling}

In literature, the study of competitive team sports may be seen to
lie between two primary goals. The first goal is to design models that make good predictions for future
match outcome. The second goal is to understand the key factors that influence
the match outcome, mostly through retrospective analysis~\citep{pollard1986home,rue2000prediction}.
As explained above, these two aspects are intrinsically connected,
and in our view they are the two
facets of a single problem: on one hand, proposed influential factors
are only scientifically valid if confirmed by falsifiable
experiments such as predictions on future matches. If the predictive
performance does not increase when information about such factors
enters the model, one should conclude by Occam's razor that these factors are actually
irrelevant\footnote{... to distinguish/characterize the observations, which in some cases may
plausibly pertain to restrictions in set of observations, rather than to causative relevance.
Hypothetical example: age of football players may be identified as unimportant for the outcome -
which may plausibly be due to the fact that the data contained no players of ages 5 or 80, say,
as opposed to player age being unimportant in general. Rephrased, it is only unimportant for cases
that are plausible to be found in the data set in the first place.}.
On the other hand, it is plausible to assume that
predictions are improved by making use of relevant factors (also known
as ``features'') become available, for example because they
are capable of explaining unmodelled random effects (noise). In light
of this, the main problem considered in this work is the
(validable and falsifiable) \textit{prediction} problem, which
in machine learning terminology is also known as the supervised learning task.

\subsection{Main questions and challenges in competitive sports outcomes prediction\label{sec:Questions}}

Given the above discussion, the major challenges may be stated as follows:\\

On the {\bf methodological} side, what are suitable models
for competitive sports outcomes? Current models are not at the same time
interpretable, easily computable, allow to use feature information on the teams/players,
and allow to predict scores or ternary outcomes.
It is an open question how to achieve this in the best way, and this manuscript
attempts to highlight a possible path.

The main technical difficulty lies in the fact that off-shelf methods do not apply
due to the structured nature of the data:
unlike in individual sports such as running and swimming where the outcome
depends only on the given team, and where the prediction task may
be dealt with classical statistics and machine learning technology
(see~\citep{blythe2015prediction} for a discussion of this in the context of running),
in competitive team sports the outcome may
be determined by potentially complex interactions between two opposing teams.
In particular, the performance of any team is not measured directly using a simple metric,
but only in relation to the opposing team's performance.\\

On the side of {\bf domain applications}, which in this manuscript is premier league football,
it is of great interest to determine the relevant factors determining the outcome,
the best way to predict, and which ranking systems are fair and appropriate.

All these questions are related to predictive modelling, as well as the availability of
suitable amounts of quality data. Unfortunately, the scarcity of features available
in systematic presentation places a hurdle to academic research in competitive team sports, especially
when it comes to assessing important factors such as team member characteristics,
or strategic considerations during the match.

Moreover, closely linked is also the question to which extent the outcomes are determined by
``chance'' as opposed to ``skill''. Since if on one hypothetical extreme, results would prove to be completely
unpredictable, there would be no empirical evidence to distinguish the matches from a game of chance
such as flipping a coin. On the other hand, importance of a measurement for predicting would strongly
suggest its importance for winning (or losing), though without an experiment not necessarily a causative link.

We attempt to address these questions in the case of premier league football within the confines of readily available data.

%\newpage
\subsection{Main contributions}

Our main contributions in this manuscript are the following:
\begin{itemize}
\item[(i)] We give what we believe to be the first comprehensive {\bf literature review} of state-of-art competitive sports modelling that comprises the multiple communities (Bradley-Terry models, \'{E}l\H{o} type models, Bayesian models, machine learning) in which research so far has been conducted mostly separately.
\item[(ii)] We present a {\bf unified Bradley-Terry-\'{E}l\H{o} model} which combines the statistical rigour of the Bradley-Terry models with fitting and update strategies similar to that found in the \'{E}l\H{o} system. Mathematically only a small step, this joint view is essential in a predictive/supervised setting as it allows efficient training and application in an on-line learning situation. Practically, this step solves some problems of the \'{E}l\H{o} system (including ranking initialization and choice of K-factor), and establishes close relations to logistic regression, low-rank matrix completion, and neural networks.
\item[(iii)] This unified view on Bradley-Terry-\'{E}l\H{o} allows us to introduce classes of joint extensions, {\bf the structured log-odds models}, which unites desirable properties of the extensions found in the disjoint communities: probabilistic prediction of scores and wins/draws/losses, batch/epoch and on-line learning, as well as the possibility to incorporate features in the prediction, without having to sacrifice structural parsimony of the Bradley-Terry models, or simplicity and computational efficiency of \'{E}l\H{o}'s original approach.
\item[(iv)] We validate the practical usefulness of the structured log-odds models in synthetic experiments and in
    {\bf answering domain questions on English Premier League data}, most prominently on the importance of features, fairness of the ranking,
    as well as on the ``chance''-``skill'' divide.
\end{itemize}

\subsection{Manuscript structure}
Section~\ref{sec:Background-and-Related} gives an overview of the mathematical setting in competitive sports prediction.
Building on the technical context, Section~\ref{sub:A-brief-summary} presents a more extensive review of the literature related to the prediction problem
of competitive sports, and introduces a joint view on Bradley-Terry and \'{E}l\H{o} type models.
Section~\ref{sec:Methods} introduces the structured log-odds models, which are validated in
empirical experiments in Section~\ref{sec:Experiments}.
Our results and possible future directions for research are discussed in section~\ref{sec:Summary-and-Conclusion}.

\subsection*{Authors' contributions}
This manuscript is based on ZQ's MSc thesis, submitted September 2016 at University College London, written under supervision of FK.
FK provided the ideas of re-interpretation and possible extensions of the \'{E}l\H{o} model.
Literature overview is jointly due to ZQ an FQ, and in parts follows some very helpful pointers by I.~Kosmidis (see below).
Novel technical ideas in Sections~\ref{sub:Extensions-of-structured} to~\ref{sub:Regularized-log-odds-matrix},
and experiments (set-up and implementation) are mostly due to ZQ.

The present manuscript is a substantial re-working of the thesis manuscript, jointly done by FK and ZQ.

\subsection*{Acknowledgements}
We are thankful to Ioannis Kosmidis for comments on an earlier form of the manuscript,
for pointing out some earlier occurrences of ideas presented in it but not given proper credit,
as well as relevant literature in the ``Bradley-Terry'' branch.

\newpage
\section{The Mathematical-Statistical Setting\label{sec:Background-and-Related}}

This section formulates the prediction task in competitive sports and fixes notation,
considering as an instance of supervised learning with several non-standard structural aspects being of relevance.

\subsection{Supervised prediction of competitive outcomes\label{sub:The-challenges-of}}

We introduce the mathematical setting for outcome prediction in competitive team sports.
As outlined in the introductory Section \ref{sub:intro_one}, three crucial features need to be taken into account in this setting:
\begin{itemize}
\item[(i)] The outcome of a pairing cannot be exactly predicted prior to the game, even with perfect knowledge of all determinates.
Hence it is preferable to predict a \textit{probabilistic} estimate for all possible match outcomes (win/draw/loss) rather than \textit{deterministically}
choosing one of them.
\item[(ii)] In a pairing, two teams play against each other, one as a home team and the other as the away or guest team. Not all pairs may play against each other, while others may play multiple times. As a mathematically prototypical (though inaccurate) sub-case one may consider all pairs playing exactly once, which gives the observations an implicit \textit{matrix structure} (row = home team, column = away team). Outcome labels and features crucially depend on the teams constituting the pairing.
\item[(iii)] Pairings take place over time, and the expected outcomes are plausibly expected to change with (possibly hidden) characteristics of the teams.
Hence we will model the \textit{temporal dependence} explicitly to be able to take it into account when building and checking predictive strategies.
\end{itemize}

\subsubsection{The Generative Model.}
Following the above discussion, we will fix a generative model as follows:
as in the standard supervised learning setting, we will consider a generative joint random variable $(X,Y)$ taking values in
$\calX\times \calY$, where $\calX$ is the set of features (or covariates, independent variables) for each \emph{pairing}, while $\calY$ is the set of labels (or outcome variables, dependent variables).

In our setting, we will consider only the cases $\calY = \{\text{win},\,\text{lose}\}$ and $\calY = \{\text{win},\,\text{lose},\,\text{draw}\}$,
in which case an observation from $\calY$ is a so-called \emph{match outcome}, as well as the case $\calY = \NN^2$, in which case an observation
is a so-called \emph{final score} (in which case, by convention, the first component of $\calY$ is of the home team), or the case of
\emph{score differences} where $\calY = \NN$ (in which case, by convention, a positive number is in favour of the home team).
From the official rule set of a game (such as football), the match outcome is uniquely determined by a score or score difference.
As all the above sets $\calY$ are discrete, predicting $\calY$ will amount to supervised \emph{classification}
(the score difference problem may be phrased as a regression problem, but we will abstain from doing so for technical reasons that become apparent later).

The random variable $X$ and its domain $\calX$ shall include information on the teams playing, as well as on the time of the match.

We will suppose there is a set $\calI$ of teams, and for $i,j\in \calI$ we will denote by $(X_{ij},Y_{ij})$ the random variable $(X,Y)$
conditioned on the knowledge that $i$ is the home team, and $j$ is the away team.
Note that information in $X_{ij}$ can include any knowledge on either single team $i$ or $j$, but also information corresponding uniquely to the pairing $(i,j)$.

We will assume that there are $Q:=\# \calI$ teams, which means that the $X_{ij}$ and $Y_{ij}$ may be arranged in $(Q\times Q)$ matrices each.

Further there will be a set $\calT$ of time points at which matches are observed. For $t\in \calT$ we will denote by $(X(t),Y(t))$ or $(X_{ij}(t),Y_{ij}(t))$
an additional conditioning that the outcome is observed at time point $t$.

Note that the indexing $X_{ij}(t)$ and $Y_{ij}(t)$ formally amounts to a double conditioning and could be written as $X|I = i, J = j, T = t$ and $Y|I = i, J = j, T = t$, where $I,J,T$ are random variables denoting the home team, the away team, and the time of the pairing. Though we do believe that the index/bracket notation is easier to carry through and to follow (including an explicit mirroring of the the ``matrix structure'') than the conditional or ``graphical models'' type notation, which is our main reason for adopting the former and not the latter.

\subsubsection{The Observation Model.}
By construction, the generative random variable $(X,Y)$ contains all information on having any pairing playing at any time,
However, observations in practice will concern two teams playing at a certain time,
hence observations in practice will only include independent samples of $(X_{ij}(t),Y_{ij}(t))$ for some $i,j\in \calI, t\in \calT$, and never full observations of $(X,Y)$ which can be interpreted as a latent variable.

Note that the observations can be, in-principle, correlated (or unconditionally dependent) if the pairing $(i,j)$ or the time $t$ is not made explicit (by conditioning which is implicit in the indices $i,j,t$).

An important aspect of our observation model will be that whenever a value of $X_{ij}(t)$ or $Y_{ij}(t)$ is observed, it will always come together with the information of the playing teams $(i,j)\in\calI^2$ and the time $t\in\calT$ at which it was observed. This fact will be implicitly made use of in description of algorithms and validation methodology.
(formally this could be achieved by explicitly exhibiting/adding $\calI\times \calI \times \calT$ as a Cartesian factor of the sampling domains $\calX$ or $\calY$ which we will not do for reasons of clarity and readability)

Two independent batches of data will be observed in the exposition. We will consider:
\begin{align*}
\mbox{a training set}\;\calD &:= \{(X^{(1)}_{i_1j_1}(t_1),Y^{(1)}_{i_1j_1}(t_1)),\dots,(X^{(N)}_{i_Nj_N}(t_N),Y^{(N)}_{i_Nj_N}(t_N))\}\\
\mbox{a test set}\;\calT &:= \{(X^{(1*)}_{i^*_1j^*_1}(t^*_1),Y^{(1*)}_{i^*_1j^*_1}(t^*_1)),\dots,(X^{(M*)}_{i^*_Mj^*_M}(t^*_M),Y^{(M*)}_{i^*_Mj^*_M}(t^*_M))\}\\
\end{align*}
where $(X^{(i)},Y^{(i)})$ and $(X^{(i*)},Y^{(i*)})$ are i.i.d.~samples from $(X,Y)$.%, and where we will denote the sets of pairing/time indices as
%\begin{align*}
%\calQ(\calD) &:=\{(i,j,t)\in \calI^2\times \calT\;:\;(i,j,t) = (i_k,j_k,t_k)\;\mbox{for some}\;k\}\\
%\calQ(\calT) &:=\{(i,j,t)\in \calI^2\times \calT\;:\;(i,j,t) = (i^*_k,j^*_k,t^*_k)\;\mbox{for some}\;k\}.
%\end{align*}
%As said in-text above, $\calQ(\calD)$ and $\calQ(\calT)$ will be considered part of the observation.

Note that unfortunately (from a notational perspective), one cannot omit the superscripts $\kappa$ as in $X^{(\kappa)}$ when defining the samples, since the figurative ``dies'' should be cast anew for each pairing taking place. In particular, if all games would consist of a single pair of teams playing where the results are independent of time, they would all be the same (and not only identically distributed) without the super-index, i.e., without distinguishing different games as different samples from $(X,Y)$.

\subsubsection{The Learning Task.} As set out in the beginning, the main task we will be concerned with is predicting future outcomes given past outcomes and features, observed from the process above. In this work, the features will be assumed to change over time slowly. It is \emph{not} our primary goal to identify the hidden features in $(X,Y)$, as they are never observed and hence not accessible as ground truth which can validate our models. However, these will be of secondary interest and considered empirically validated by a well-predicting model.

More precisely, we will describe methodology for learning and validating predictive models of the type
$$f: \calX\times \calI \times \calI \times \calT \rightarrow \Distr (\calY),$$
where $\Distr (\calY)$ is the set of (discrete probability) distributions on $\calY$.
That is, given a pairing $(i,j)$ and a time point $t$ at which the teams $i$ and $j$ play, and information of type $x=X_{ij}(t)$, make a probabilistic prediction $f(x,i,j,t)$ of the outcome.

Most algorithms we discuss will \emph{not} use added information in $\calX$, hence will be of type $f:\calI \times \calI \times \calT \rightarrow \Distr (\calY)$. Some will disregard the time in $\calT$. Indeed, the latter algorithms are to be considered scientific baselines above which any algorithm using information in $\calX$ and/or $\calT$ has to improve.

The models $f$ above will be learnt on a training set $\calD$, and validated on an independent test set $\calT$ as defined above.
In this scenario, $f$ will be a random variable which may implicitly depend on $\calD$ but will be independent of $\calT$.
The learning strategy - which is $f$ depending on $\calD$ - may take any form and is considered in a full black-box sense.
In the exposition, it will in fact take the form of various parametric and non-parametric prediction algorithms.

The goodness of such an $f$ will be evaluated by a loss
$L:\Distr (\calY)\times \calY\rightarrow \RR$ which compares a probabilistic prediction to the true observation.
The best $f$ will have a small expected generalization loss
$$\varepsilon (f|i,j,t) := \EE_{(X,Y)}\left[L\left(f(X_{ij}(t),i,j,t),Y_{ij}(t)\right)\right]$$
at any future time point $t$ and for any pairing $i,j$.
Under mild assumptions, we will argue below that this quantity is estimable from $\calT$ and only mildly dependent on $t,i,j$.

Though a good form for $L$ is not a-priori clear. Also, it is unclear under which assumptions $\varepsilon (f|t)$ is estimable, due do the conditioning on $(i,j,t)$ in the training set. These special aspects of the competitive sports prediction settings will be addressed in the subsequent sections.

\subsection{Losses for probablistic classification}

In order to evaluate different models, we need a criterion to measure
the goodness of probabilistic predictions. The most common error metric
used in supervised classification problems is the prediction accuracy.
However, the accuracy is often insensitive to probabilistic predictions.

For example, on a certain test case model A predicts a win probability of 60\%, while model B predicts a win probability of 95\%. If the
actual outcome is not win, both models are wrong. In terms of prediction
accuracy (or any other non-probabilistic metric), they are equally wrong because both of them made
one mistake. However, model B should be considered better than model A since it predicted the ``true'' outcome with higher accuracy.

Similarly, if a large number of outcomes of a fair coin toss have been observed as training data, a model that predicts 50\% percent for
both outcomes on any test data point should be considered more accurate than a model that predicts 100\% percent for either outcome 50\% of the time.

There exists two commonly used criteria that take into account the probabilistic nature of predictions which we adopt. The first one is the Brier score (Equation~\ref{eq:BS} below)
and the second is the log-loss or log-likelihood loss (Equation~\ref{eq:oob_log} below).
Both losses compare a distribution to an observation, hence mathematically have the signature of a function $\Distr (\calY)\times \calY\rightarrow \RR$.
By (very slight) abuse of notation, we will identify distributions on (discrete) $\calY$ with its probability mass function; for a distribution $p$, for $y\in \calY$ we write $p_y$ for mass on the observation $y$ (= the probability to observe $y$ in a random experiment following $p$).

With this convention, log-loss $L_\ell$ and Brier loss $L_{\text{Br}}$ are defined as follows:

\begin{eqnarray}
L_\ell:& (p,y)\mapsto& - \log p_y \label{eq:BS}\\
L_{\text{Br}}:& (p,y)\mapsto& (1-p_y)^2 + \sum_{y\in \calY\setminus \{y\}} p_y^2\label{eq:oob_log}
\end{eqnarray}

The log-loss and the Brier loss functions have the following properties:
\begin{enumerate}
\item[(i)] the Brier Score is only defined on $\calY$ with an addition/subtraction and a norm defined.
This is not necessarily the case in our setting where it may be that $\calY = \{\text{win},\,\text{lose},\,\text{draw}\}$.
In literature, this is often identified with $\calY = \{1,0,-1\}$, though this identification is arbitrary, and the Brier score may change depending on which numbers are used.

On the other hand, the log-loss is defined for any $\calY$ and remains unchanged under any renaming or renumbering of a discrete $\calY$.

\item[(ii)] For a joint random variable $(X,Y)$ taking values in $\calX\times \calY$, it can be shown that the expected losses $\EE\left[ L_\ell(f(X),Y) \right]$ are minimized by the ``correct'' prediction $f: x\mapsto \left(p_y = P(Y=y|X=x)\right)_{y\in \calY}$.
\end{enumerate}

The two loss functions usually are introduced as empirical losses on a test set $\calT$, i.e.,
$$\widehat{\varepsilon}_\calT(f) = \frac{1}{\# \calT}\sum_{(x,y)\in \calT} L_*(x,y).$$
The empirical log-loss is the (negative log-)likelihood of the test predictions.

The empirical Brier loss, usually called the ``Brier score'', is a straightforward translation of the mean squared error
used in regression problems to the classification setting, as the expected
mean squared error of predicted confidence scores.
However, in certain cases, the Brier score is hard to interpret and may behave in
unintuitive ways~\citep{jewson2004problem}, which may partly be seen as a phenomenon caused
by above-mentioned lack of invariance under class re-labelling.

Given this and the interpretability of the empirical log-loss as a likelihood,
we will use the log-loss as principal evaluation
metric in the competitive outcome prediction setting.

\subsection{Learning with structured and sequential data\label{sub:Working-with-sequential}}

The dependency of the observed data on pairing and time makes the prediction task at hand non-standard.
We outline the major consequences for learning and model validation, as well as the implicit assumptions which allow us to tackle these.
We will do this separately for the pairing and the temporal structure, as these behave slightly differently.

\subsubsection{Conditioning on the pairing}
Match outcomes are observed for given pairings $(i,j)$, that is, each feature-label-pair will be of form $(X_{ij},Y_{ij})$, where as above the subscripts denote conditioning on the pairing. Multiple pairings may be observed in the training set, but not all; some pairings may never be observed.

This has consequences for both learning and validating models.\\

For {\bf model learning}, it needs to be made sure that the pairings to be predicted \emph{can} be predicted from the pairings observed. With other words, the label $Y^*_{ij}$ in the test set that we want to predict is (in a practically substantial way) dependent on the training set $\calD = \{(X^{(1)}_{i_1j_1},Y^{(1)}_{i_1j_1}),\dots,(X^{(N)}_{i_Nj_N},Y^{(N)}_{i_Nj_N}) \}$. Note that smart models will be able to predict the outcome of a pairing even if it has not been observed before, and even if it has, it will use information from other pairings to improve its predictions

For various parametric models, ``predictability'' can be related to completability of a data matrix with $Y_{ij}$ as entries. In section~\ref{sec:Methods},
we will relate \'{E}l\H{o} type models to low-rank matrix completion algorithms; completion can be understood as low-rank completion,
hence predictability corresponds to completability. Though, exactly working completability out is not the main is not the primary aim of this manuscript,
and for our data of interest, the English Premier League, all pairings are observed in any given year, so completability is not an issue.
Hence we refer to~\cite{kiraly2015algebraic} for a study of low-rank matrix completability. General parametric models may be treated along similar lines.\\

For model-agnostic {\bf model validation}, it should hold that the expected generalization loss
$$\varepsilon (f|i,j) := \EE_{(X,Y)}\left[L\left(f(X_{ij},i,j),Y_{ij}\right)\right]$$
can be well-estimated by empirical estimation on the test data. For league level team sports data sets, this can be achieved by having multiple years of data available.
Since even if not all pairings are observed, usually the set of pairings which \emph{is} observed is (almost) the same in each year, hence the pairings will be similar in the training and test set if whole years (or half-seasons) are included.
Further we will consider an average over all observed pairings, i.e., we will compute the empirical loss on the training set $\calT$ as
$$\widehat {\varepsilon} (f) := \frac{1}{\# \calT}\sum_{(X_{ij},Y_{ij})\in \calT} L\left(f(X_{ij},i,j),Y_{ij}\right)$$
By the above argument, the set of all observed pairings in any given year is plausibly modelled as similar, hence it is plausible to conclude that this empirical loss estimates some expected generalization loss
$$\varepsilon(f) := \EE_{X,Y,I,J}\left[L\left(f(X_{IJ},I,J),Y_{IJ}\right)\right]$$
where $I,J$ (possibly dependent) are random variables that select teams which are paired.

Note that this type of aggregate evaluation does not exclude the possibility that predictions for single teams (e.g., newcomers or after re-structuring) may be inaccurate, but only that the ``average'' prediction is good. Further, the assumption itself may be violated if the whole league changes between training and test set.

\subsubsection{Conditioning on time}
As a second complication, match outcome data is gathered through time. The data
set might display temporal structure and correlation with time. Again, this has consequences for learning and validating the models.\\

For {\bf model learning}, models should be able to intrinsically take into
account the temporal structure - though as a baseline, time-agnostic models should be tried.
A common approach for statistical models is to assume a temporal structure in the latent
variables that determine a team's strength. A different and somewhat
ad-hoc approach proposed by \citet{dixon1997modelling} is to assign
lower weights to earlier observations and estimate parameter by maximizing
the weighted log-likelihood function. For machine learning models,
the temporal structure is often encoded with handcrafted features.

Similarly, one may opt to choose a model that can be updated as time progresses.
A common ad-hoc solution is to re-train the model after a certain amount
of time (a week, a month, etc), possibly with temporal discounting, though there is no general consensus about
how frequently the retraining should be performed.
Further there are genuinely updating models, so-called on-line learning models, which update model
parameters after each new match outcome is revealed.\\

For {\bf model evaluation}, the sequential nature of the data poses a severe restriction:
Any two data points were measured at certain time points, and one can not assume that they are not correlated through time information.
That such correlation exists is quite plausible in the domain application, as a team would be expected to perform more similarly at close time points than at distant time points.
Also, we would like to make sure that we fairly test the models for their prediction accuracy -
hence the validation experiment needs to mimic the ``real world'' prediction process, in which the predicted outcomes will be in the temporal future of the training data.
Hence the test set, in a validation experiment that should quantify goodness of such prediction, also needs to be in the temporal future of the training set.

In particular, the common independence assumption that allows application of re-sampling strategies such as the K-fold cross-validation method~\citep{stone1974cross},
which guarantees the expected loss to be estimated by the empirical loss, is violated. In the presence of temporal correlation,
the variance of the error metric may be underestimated, and  the error metric itself will, in general, be mis-estimated.
Moreover, the validation method
will need to accommodate the fact that the model may be updated on-line
during testing. In literature, model-independent validation strategies for data
with temporal structure is largely an unexplored (since technically difficult) area. Nevertheless, developing
a reasonable validation method is crucial for scientific model assessment. A plausible
validation method is introduced in section
\ref{sub:Tunning-and-validation} in detail.
It follows similar lines as the often-seen ``temporal cross-validation'' where training/test splits are always temporal, i.e., the training data points are in the temporal past of the test data points, for multiple splits. An earlier occurrence of such a validation strategy may be found in~\cite{hyndman2014forecasting}.

This strategy comes without strong estimation guarantees and is part heuristic; the empirical loss will estimate the generalization loss as long as statistical properties do not change as time shifts forward, for example under stationarity assumptions. While this implicit assumption may be plausible for the English Premier League, this condition is routinely violated in financial time series, for example.

\newpage
\section{Approaches to competitive sports prediction\label{sub:A-brief-summary}}

In this section, we give a brief overview over the major approaches to prediction in competitive sports found in literature. Briefly, these are:
\begin{enumerate}
\item[(a)] The Bradley-Terry models and extensions.
\item[(b)] The \'{E}l\H{o} model and extensions.
\item[(c)] Bayesian models, especially latent variable models and/or graphical models for the outcome and score distribution.
\item[(d)] Supervised machine learning type models that use domain features for prediction.
\end{enumerate}

(a) The {\bf Bradley-Terry} model is the most influential statistical approach to ranking based on competitive
observations~\citep{bradley1952rank}.
With its original applications in psychometrics, the goal of the class of Bradley-Terry models is to
estimate a hypothesized rank or skill level from observations of pairwise competition outcomes (win/loss).
Literature in this branch of research is, usually, primarily concerned not with prediction, but estimation of
a ``true'' rank or skill, existence of which is hypothesized, though prediction
of (binary) outcome probabilities or odds is well possible within the paradigm.
A notable exception is the work of~\cite{stanescu2011rating} where the problem is in essence formulated
as supervised prediction, similar to our work.
Mathematically, Bradley-Terry models may be seen as log-linear two-factor models that, at the state-of-art are usually
estimated by (analytic or semi-analytic) likelihood maximization~\citep{hunter2004mm}.
Recent work has seen many extensions of the Bradley-Terry models, most notably for modelling of ties~\cite{rao1967ties},
making use of features~\cite{firth2012bradley} or for explicit modelling the time dependency of skill~\cite{cattelan2013dynamic}.\\

(b) The {\bf \'{E}l\H{o} system} is one of the earliest attempts to model competitive sports
and, due to its mathematical simplicity, well-known and widely-used by practitioners~\citep{elo1978rating}.
Historically, the \'{E}l\H{o} system is used for chess rankings, to assign a rank score to chess players.
Mathematically, the \'{E}l\H{o} system only uses information about the historical match outcomes. The \'{E}l\H{o}
system assigns to each team a parameter, the so-called \'{E}l\H{o} rating.
The rating reflects a team's competitive skills: the team with higher
rating is stronger.
As such, the \'{E}l\H{o} system is, originally, not a predictive model or a statistical model in the usual sense.
However, the \'{E}l\H{o} system also gives a probabilistic prediction for the \textit{binary} match outcome based
on the ratings of two teams.
After what appears to have been a period of parallel development that is still partly ongoing,
it has been recently noted by members of the Bradley-Terry community that the \'{E}l\H{o} prediction heuristic
is mathematically equivalent to the prediction via the simple Bradley-Terry
model~\citep[see][, section 2.1]{coulom2007computing}.\\
The \'{E}l\H{o} ratings are learnt via an update rule that is applied whenever a new outcome is observed.
This suggested update strategy is inherently algorithmic and later shown to be closely related to
on-line learning strategies in neural network; to our knowledge it appears first in \'{E}l\H{o}'s work
and is not found in the Bradley-Terry strain.\\

(c) The {\bf Bayesian paradigm} offers a natural framework to model match outcomes probabilistically,
and to obtain probabilistic predictions as the posterior predictive distribution.
Bayesian parametric models also allow researchers to inject expert
knowledge through the prior distribution. The prediction function
is naturally given by the posterior distribution of the scores, which
can be updated as more observations become available.

Often, such models explicitly model not only the outcome but also the score distribution,
such as Maher's model~\cite{maher1982modelling} which models outcome scores
based on independent Poisson random variables with team-specific means.
\citet{dixon1997modelling}
extend Maher's model by introducing a correlation effect between
the two final scores.
More recent models also include dynamic components to model
temporal dependence~\citep{glickman1998state,rue2000prediction,crowder2002dynamic}.
Most models of this type only use historical match outcomes as features,
see \citet{constantinou2012pi} for an exception.\\

(d) More recently, the method-agnostic {\bf supervised machine learning paradigm} has been
applied to prediction of match outcomes~\cite{liu2010beating,hucaljuk2011predicting,odachowski2012using}.
The main rationale in this branch of research is that the best model is not known, hence
a number of off-shelf predictors are tried and compared in a benchmarking experiment.
Further, these models are able to make use of features other than previous outcomes easily.
However, usually, the machine learning  models are trained in-batch, i.e., not following a dynamic update or on-line learning strategy,
and they need to be re-trained periodically to incorporate new observations.\\

In this manuscript, we will re-interpret the \'{E}l\H{o} model and its update rule as
the simplest case of a structured extension of predictive logistic (or generalized linear) regression models, and the canonical gradient ascent update of its likelihood
- hence, in fact, giving it a parametric form not entirely unlike the models mentioned in (b),
In the subsequent sections, this will allow us to complement it with the beneficial properties of the machine learning approach (c),
most notably the addition of possibly complex features, paired with the \'{E}l\H{o} update rule which can be shown generalize to an on-line update strategy.

More detailed literature and technical overview is given given in the subsequent sections.
The \'{E}l\H{o} model and its extensions, as well as its novel parametric interpretation, are reviewed in Section~\ref{sub:The-Elo-model}.
Section \ref{sub:Statistical-and-latent}
reviews other parametric models for predicting final scores. Section \ref{sub:Feature-based-machine-learning} reviews the use of
machine learning predictors and feature engineering for sports prediction.

\subsection{The Bradley-Terry-\'{E}l\H{o} models\label{sub:The-Elo-model}}

This section reviews the Bradley-Terry models, the \'{E}l\H{o} system, and closely related variants.

We give the above-mentioned joint formulation, following the modern rationale of considering as a ``model'' not only a generative specification, but also algorithms for training, predicting and updating its parameters.
As the first seems to originate with the work of~\cite{bradley1952rank}, and the second in the on-line update heuristic of~\cite{elo1978rating},
we argue that for giving proper credit, it is probably more appropriate to talk about Bradley-Terry-\'{E}l\H{o} models
(except in the specific hypothesis testing scenario covered in the original work of Bradley and Terry).

Later, we will attempt to understand the \'{E}l\H{o} system as an on-line update of a structured logistic odds model.

\subsubsection{The original formulation of the \'{E}l\H{o} model}

We will first introduce the original version of the \'{E}l\H{o} model, following~\citep{elo1978rating}.
As stated above, its original form which is still applied for determining the official chess ratings (with minor domain-specific modifications),
is neither a statistical model nor a predictive model in the usual sense.

Instead, the original version is centered around the ratings $\theta_i$ for each team $i$.
These ratings are updated via the \'{E}l\H{o} model rule, which we explain (for sake of clarity) for the case of no draws:
After observing a match between (home) team $i$ and (away) team $j$, the ratings of teams $i$ and $j$ are updated as

\begin{eqnarray}
\theta_{i} & \leftarrow&\theta_{i}+K\left[S_{ij}-p_{ij}\right]\label{eq:elo_update}\\
\theta_{j} & \leftarrow&\theta_{j}-K\left[S_{ij}-p_{ij}\right]\nonumber
\end{eqnarray}

where $K$, often called ``the K factor'', is an arbitrarily chosen constant, that is, a model parameter usually set per hand.
$S_{ij}$ is $1$ if team/player $i$ has been observed to win, and $0$ otherwise.

Further, $p_{ij}$ is the probability of $i$ winning against $j$
which is predicted from the ratings prior to the update by

\begin{equation}
p_{ij}=\sigma(\theta_{i}-\theta_{j})\label{eq:elo_prob}
\end{equation}
where $\sigma: x\mapsto  \left(1+\exp(-x)\right)^{-1}$ is the logistic function (which has a sigmoid shape, hence is also often called ``the sigmoid'').
Sometimes a home team parameter $h$ is added to account for home advantage, and the predictive equation becomes
\begin{equation}
p_{ij}=\sigma(\theta_{i}-\theta_{j} + h)\label{eq:Elo_gen}
\end{equation}

\'{E}l\H{o}'s update rule (Equation~\ref{eq:elo_update}) makes sense intuitively because the term $(S_{ij}-p_{ij})$
can be thought of as the discrepancy between what is expected, $p_{ij}$,
and what is observed, $S_{ij}$. The update will be larger if the
current parameter setting produces a large discrepancy. However, a concise
theoretical justification has not been articulated in literature.
If fact, \'{E}l\H{o} himself commented that ``the logic of the equation
is evident without algebraic demonstration''~\citep{elo1978rating}
- which may be true in his case, but not satisfactory
in an applied scientific nor a theoretical/mathematical sense.

As an initial issue, it has been noted that the whole model is invariant under joint re-scaling of the $\theta_i$, and the parameters $K,h$, as well as under arbitrary choice of zero for the $\theta_i$ (i.e., adding of a fixed constant $c\in\RR$ to all $\theta_i$).
Hence, fixed domain models will usually choose zero and scale arbitrarily. In chess rankings, for example, the
formula includes additional scaling constants of the form $p_{ij}=\left(1+10^{-(\theta_{i}-\theta_{j})/400}\right)^{-1}$;
scale and zero are set through fixing some historical chess players'
rating, which happens to set the ``interesting'' range in the positive thousands\footnote{A common misunderstanding here is that no \'{E}l\H{o} ratings below zero may occur.
This is, in-principle, wrong, though it may be extremely unlikely in practice if the arbitrarily chosen zero is chosen low enough.}.
One can show that there are no more parameter redundancies, hence scaling/zeroing turns out not to be a problem if kept in mind.

However, three issues are left open in this formulation:
\begin{enumerate}
\item[(i)] How the ratings for players/teams are determined who have never played a game before.
\item[(ii)] The choice of the constant/parameter $K$, the ``K-factor''.
\item[(iii)] If a home parameter $h$ is present, its size.
\end{enumerate}

These issues are usually addressed in everyday practice by (more or less well-justified) heuristics.

The parametric and probabilistic supervised setting in the following sections yields more principled ways to address this.
step (i) will become unnecessary by pointing out a batch learning method;
the constant $K$ in (ii) will turn out to be the learning rate in a gradient update,
hence it can be cross-validated or entirely replaced by a different strategy for learning the model.
Parameters such as $h$ in (iii) will be interpretable as a logistic regression coefficient.

See for this the discussions in Sections~\ref{sub:Training-structured-log-odds},~\ref{sub:Training-structured-log-odds.batch} for (i),(ii), and Section~\ref{sec:specialcases} for (iii).

\subsubsection{Bradley-Terry-\'{E}l\H{o} models\label{sub:The-probabilistic-interpretation}}

As outlined in the initial discussion, the class of Bradley-Terry models introduced by~\citep{bradley1952rank} may be interpreted as a
proper statistical model formulation of the \'{E}l\H{o} prediction heuristic.

Despite their close mathematical vicinity, it should be noted that classically Bradley-Terry and \'{E}l\H{o} models are usually applied and interpreted differently, and consequently fitted/learnt differently: while both models estimate a rank or score, the primary (historical) purpose of the Bradley-Terry is to estimate the rank, while the \'{E}l\H{o} system is additionally intended to supply easy-to-compute updates as new outcomes are observed, a feature for which it has historically paid for by lack of mathematical rigour.
The \'{E}l\H{o} system is often invoked to predict future outcome probabilities, while the Bradley-Terry models usually do not see predictive use
(despite their capability to do so, and the mathematical equivalence of both predictive rules).

However, as mentioned above and as noted for example by~\citep{coulom2007computing}, a joint mathematical formulation can be found, and as we will show,
the different methods of training the model may be interpreted as variants of likelihood-based batch or on-line strategies.

%The first proper statistical formulation of the \'{E}l\H{o} model was given by~\citet{glickman1995comprehensive}.
%However, since Glickman's model makes unnecessary (and partly unidentifiable) assumptions, we will discuss it in the subsequent section.
%We introduce a simpler parametric formulation instead, which is more natural in the on-line learning setting and lends itself more straightforwardly to extensions.
%Here, we consider only the case of a binary (home team win/lose = 1/0) outcome; extensions and empirical validation will be given in later sections.

The parametric formulation is quite similar to logistic regression models, or generalized linear models,
in that we will use a link function and define a model for the outcome odds.
Recall, the odds for a probability $p$ are $\odds(p) := p/(1-p)$, and the logit function is $\logit: x\mapsto \log\odds(x) = \log x - \log(1-x)$
(sometimes also called the ``log-odds function'' for obvious reasons).
A straightforward calculation shows that $\logit^{-1} = \sigma$, or equivalently, $\sigma(\logit(x)) = x$ for any $x$, i.e., the logistic function is the inverse of the logit (and vice versa $\logit(\sigma(x)) = x$ by the symmetry theorem for the inverse function).

Hence we can posit the following two equivalent equations in latent parameters $\theta_i$ as \emph{definition} of a predictive model:
\begin{eqnarray}
p_{ij} & = &\sigma(\theta_{i}-\theta_{j}) \label{eq:elo_prob2}\\
\logit(p_{ij})& = &\theta_{i}-\theta_{j}\nonumber
\end{eqnarray}
That is, $p_{ij}$ in the first equation is interpreted as a predictive probability; i.e., $Y_{ij}\sim \mbox{Bernoulli} (p_{ij})$.
The second equation interprets this prediction in terms of a generalized linear model with a response function that is linear in the $\theta_i$.
We will write $\theta$ for the vector of $\theta_i$; hence the second equation could also be written, in vector notation,
as $\logit(p_{ij}) = \left\langle e_i - e_j, \theta \right\rangle$. Hence, in particular, the matrix with entries $\logit(p_{ij})$ has rank (at most) two.

Fitting the above model means estimating its latent variables $\theta$.
This may be done by considering the \emph{likelihood} of the latent parameters $\theta_i$ given the training data.
For a single observed match outcome $Y_{ij}$, the log-likelihood of $\theta_i$ and $\theta_j$ is
$$\ell (\theta_i,\theta_j|Y_{ij}) = Y_{ij}\log (p_{ij}) + (1-Y_{ij})\log (1-p_{ij}),$$
where the $p_{ij}$ on the right hand side need to be interpreted as functions of $\theta_i,\theta_j$ (namely, as in equation~\ref{eq:elo_prob2}).
We call $\ell (\theta_i,\theta_j|Y_{ij})$ the \emph{one-outcome} log-likelihood as it is based on a single data point.
Similarly, if multiple training outcomes $\calD = \{Y_{i_1j_1}^{(1)},\dots,Y_{i_Nj_N}^{(N)}\}$ are observed, the log-likelihood of the vector $\theta$ is
$$\ell (\theta|\calD) = \sum_{k=1}^N \left[Y^{(k)}_{i_kj_k}\log (p_{i_kj_k}) + (1-Y_{i_kj_k}^{(k)})\log (1-p_{i_kj_k})\right]$$
We will call $\ell (\theta|\calD)$ the \emph{batch log-likelihood} as the training set contains more than one data point.

The derivative of the one-outcome log-likelihood is
$$\frac{\partial}{\partial \theta_i} \ell (\theta_i,\theta_j|Y_{ij}) = Y_{ij} (1- p_{ij}) - (1-Y_{ij}) p_{ij} = Y_{ij} - p_{ij},$$
hence the $K$ in the \'{E}l\H{o} update rule (see equation~\ref{eq:elo_update}) may be updated as a gradient ascent rate or learning coefficient in an on-line likelihood update.
We also obtain a batch gradient from the batch log-likelihood:
$$\frac{\partial}{\partial \theta_i} \ell (\theta|\calD) = \left[Q_{i} - \sum_{(i,j)\in G_i} p_{ij}\right],$$
where, $Q_{i}$ is team $i$'s number of wins minus number of losses observed in $\calD$, and $G_i$ is the (multi-)set of (unordered) pairings team $i$ has participated in $\calD$.
The batch gradient directly gives rise to a batch gradient update
$$\theta_{i}\leftarrow\theta_{i}+K\cdot\left[Q_{ij}-\sum_{(i,j)\in G_i} p_{ij}\right].$$

Note that the above model highlights several novel, interconnected, and possibly so far unknown
(or at least not jointly observed) aspects of Bradley-Terry and \'{E}l\H{o} type models:
\begin{enumerate}
\item[(i)] The \'{E}l\H{o} system can be seen as a learning algorithm for a logistic odds model with latent variables, the Bradley-Terry model
(and hence, by extension, as a full fit/predict specification of a certain one-layer neural network).
\item[(ii)] The Bradley-Terry and \'{E}l\H{o} model may simultaneously be interpreted as Bernoulli observation models of a rank two matrix.
\item[(iii)] The gradient of the Bradley-Terry model's log-likelihood gives rise to a (novel) batch gradient and a single-outcome gradient ascent update.
A single iteration per-sample of the latter (with a fixed update constant) is \'{E}l\H{o}'s original update rule.
\end{enumerate}

These observations give rise to a new family of models: the structured log-odds models that will be discussed in Section~\ref{sec:Methods} and~\ref{sub:The-structured-log-odds},
together with concomitant gradient update strategies of batch and on-line type.
This joint view also makes extensions straightforward, for example, the ``home team parameter'' $h$ in the common extension $p_{ij}=\sigma(\theta_{i}-\theta_{j} + h)$
of the \'{E}l\H{o} system may be interpreted as Bradley-Terry model with an intercept term, with log-odds $\logit(p_{ij}) = \left\langle e_i - e_j, \theta \right\rangle + h$,
that is updated by the one-outcome \'{E}l\H{o} update rule.

Since more generally, the structured log-odds models arise by combining the parametric form of the Bradley-Terry model with \'{E}l\H{o}'s update strategy,
we also argue for synonymous use of the term ``Bradley-Terry-\'{E}l\H{o} models'' whenever Bradley-Terry models are updated batch, or epoch-wise,
or whenever they are, more generally, used in a predictive, supervised, or on-line setting.

\subsubsection{Glickman's Bradley-Terry-\'{E}l\H{o} model}
For sake of completeness and comparison, we discuss the probabilistic formulation of~\citet{glickman1995comprehensive}.
In this fully Bayesian take on the Bradley-Terry-\'{E}l\H{o} model, it is assumed that there is a latent random variable
$Z_{i}$ associating with team $i$. The latent variables are statistically
independent and they follow a specific generalized extreme value (GEV)
distribution:
\[
Z_{i}\sim\text{GEV}(\theta_{i},\thinspace1,\thinspace0)
\]
where the mean parameter $\theta_{i}$ varies across teams, and the
other two parameters are fixed at one and zero.
The density function of $\text{GEV}(\mu,\thinspace1,\thinspace0)$,
$\mu\in\mathbb{R}$ is
\[
p(x|\mu)=\exp\left(-(x-\mu)\right)\cdot\exp\left(-\exp\left(-(x-\mu)\right)\right)
\]

The model further assumes that team $i$ wins over team $j$ in a
match if and only if a random sample ($Z_{i}$, $Z_{j}$) from the
associated latent variables satisfies $Z_{i}>Z_{j}$.
It can be shown that the difference variables $(Z_{i}-Z_{j})$ then happen to follow a logistic
distribution with mean $\theta_{1}-\theta_{2}$ and scale parameter
1, see~\citep{resnick2013extreme}.

Hence, the (predictive) winning probability for
team $i$ is eventually given by \'{E}l\H{o}'s original equation~\ref{eq:elo_prob} which is equivalent to the Bradley-Terry-odds.
In fact, the arguably strange parametric form for the distribution $f$ of the $Z_i$
makes the impression of being chosen for this particular, singular reason.

We argue, that Glickman's model makes unnecessary assumptions
through the latent random variables $Z_i$ which furthermore carry an unnatural distribution .
This is certainly true in the frequentist interpretation, as the parametric model in Section~\ref{sub:The-probabilistic-interpretation}
is not only more parsimonious as it does not assume a process that generates the $\theta_i$,
but also it avoids to assume random variables that are never directly observed (such as the $Z_i$).
This is also true in the Bayesian interpretation, where a prior is assumed on the $\theta_i$ which then indirectly
gives rise to the outcome via the $Z_i$.

Hence, one may argue by Occam's razor, that modelling the $Z_i$ is unnecessary, and,
as we believe, may put obstacles on the path to the existing and novel extensions in Section~\ref{sec:Methods} that would otherwise appear natural.

\subsubsection{Limitations of the Bradley-Terry-\'{E}l\H{o} model and existing remedies\label{sub:Limitations-Elo}}

We point out some limitations of the original Bradley-Terry and \'{E}l\H{o} models which we attempt to address in Section~\ref{sec:Methods}.

\paragraph{Modelling draws}

The original Bradley-Terry and \'{E}l\H{o} models do not model the possibility of a draw. This
might be reasonable in official chess tournaments where players play on
until draws are resolved. However, in many competitive sports a significant
number of matches end up as a draw - for example, in the English Premier
League about twenty percent of the matches. Modelling
the possibility of draw outcome is therefore very relevant.
One of the first extensions of the Bradley-Terry model, the ternary outcome model by~\citet{rao1967ties},
was suggested to address exactly this shortcoming. The strategy
for modelling draws in the joint framework, closely following this work,
is outlined in Section~\ref{sub:Modeling-tenary-outcomes}.

\paragraph{Using final scores in the model}

The Bradley-Terry-\'{E}l\H{o} model only takes into account the binary outcome
of the match. In sports such as football, the final scores for both
teams may contain more information. Generalizations exist to tackle
this problem. One approach is adopted by the official FIFA Women’s
football ranking~\citep{FIFAwoman}, where the actual outcome of the
match is replaced by the \textquotedbl{}Actual Match Percentage\textquotedbl{},
a quantity that depends on the final scores. FiveThirtyEight, an online
media, proposed another approach~\citep{FiveThirtyEightELO}. It introduces
the ``Margin of Victory Multiplier'' in the rating system to adjust
the K-factor for different final scores.

In a survey paper,~\citet{lasek2013predictive} showed empirical evidence
that rating methods that take into account the final scores often
outperform those that do not. However, it is worth noticing that the existing
methods often rely on heuristics and their mathematical justifications
are often unpublished or unknown. We describe a principled way
to incorporate final scores in Section~\ref{sub:Using-score-difference}
into the framework, following ideas of~\citet{dixon1997modelling}.

\paragraph{Using additional features}

The Bradley-Terry-\'{E}l\H{o} model only takes into account very limited information.
Apart from previous match outcomes, the only feature it uses is the
identity of home and away teams. There are many other potentially
useful features. For example, whether the team is recently promoted
from a lower-division league, or whether a key player is absent from
the match. These features may help make better prediction if they
are properly modeled. In Section~\ref{sub:covariate}, we
extend the Bradley-Terry-\'{E}l\H{o} model to a logistic odds model
that can also make use of features, along lines similar to the feature-dependent
models of~\citet{firth2012bradley}.

\subsection{Domain-specific parametric models\label{sub:Statistical-and-latent}}

We review a number of parametric and Bayesian models that have been considered in literature to model competitive sports outcomes.
A predominant property of this branch of modelling is that the final scores are explicitly modelled.

\subsubsection{Bivariate Poisson regression and extensions}

\citet{maher1982modelling} proposed to model the final scores as
independent Poisson random variables. If team $i$ is playing at home
field against team $j$, then the final scores $S_{i}$ and $S_{j}$
follows
\begin{eqnarray*}
S_{i} & \sim & \text{Poisson}(\alpha_{i}\beta_{j}h)\\
S_{j} & \sim & \text{Poisson}(\alpha_{j}\beta_{i})
\end{eqnarray*}
where $\alpha_{i}$ and $\alpha_{j}$ measure the 'attack' rates,
and $\beta_{i}$ and $\beta_{j}$ measure the 'defense' rates of the
teams. The parameter $h$ is an adjustment term for home advantage.
The model further assumes that all historical match outcomes are independent.
The parameters are estimated from maximizing the log-likelihood function
of all historical data. Empirical evidence suggests that the Poisson
distribution fits the data well. Moreover, the Poisson distribution can
be derived as the expected number of events during a fixed time period
at a constant risk. This interpretation fits into the framework of
competitive team sports.

\citet{dixon1997modelling} proposed two modifications to Maher's
model. First, the final scores $S_{i}$ and $S_{j}$ are allowed to
be correlated when they are both less than two. The model employs
a free parameter $\rho$ to capture this effect. The joint probability
function of $S_{i},S_{j}$ is given by the bivariate Poisson distribution
\ref{eq:DC_prob}:
\begin{equation}
P(S_{i}=s_{i},S_{j}=s_{j})=\tau_{\lambda,\mu}(s_{i},s_{j})\frac{\lambda^{s_{i}}\exp(-\lambda)}{s_{i}!}\cdot\frac{\lambda^{s_{j}}\exp(-\mu)}{s_{j}!}\label{eq:DC_prob}
\end{equation}
where
\begin{eqnarray*}
\lambda & = & \alpha_{i}\beta_{j}h\\
\mu & = & \alpha_{j}\beta_{i}
\end{eqnarray*}
and
\[
\tau_{\lambda,\mu}(s_{i},s_{j})=\begin{cases}
1-\lambda\mu\rho & if\,s_{i}=s_{j}=0,\\
1+\lambda\rho & if\,s_{i}=0,\,s_{j}=1,\\
1+\mu\rho & if\,s_{i}=1,\,s_{j}=0,\\
1-\rho & if\,s_{i}=s_{j}=1,\\
1 & otherwise.
\end{cases}
\]
The function $\tau_{\lambda,\mu}$ adjusts the probability function
so that drawing becomes less likely when both scores are low. The
second modification is that the Dixon-Coles model no longer assumes
match outcomes are independent through time. The modified log-likelihood
function of all historical data is represented as a weighted sum of
log-likelihood of individual matches illustrated in equation \ref{eq:weighted_log_lik},
where $t$ represents the time index. The weights are heuristically
chosen to decay exponentially through time in order to emphasize more
recent matches.
\begin{equation}
\ell=\sum_{t=1}^{T}\exp(-\xi t)\log\left[P(S_{i}(t)=s_{i}(t),\,S_{j}(t)=s_{j}(t))\right]\label{eq:weighted_log_lik}
\end{equation}
The parameter estimation procedure is the same as Maher's model. Estimates
are obtained from batch optimization of modified log-likelihood.

\citet{karlis2003analysis} explored several other possible parametrization
of the bivariate Poisson distribution including those proposed by
~\citet{kocherlakota1992bivariate}, and \citet{johnson1997discrete}.
The authors performed a model comparison between Maher's independent
Poisson model and various bivariate Poisson models based on AIC and
BIC. However, the comparison did not include the Dixon-Coles model.
\citet{goddard2005regression} performed a more comprehensive model
comparison based on their forecasting performance.

\subsubsection{Bayesian latent variable models}

\citet{rue2000prediction} proposed a Bayesian parametric model based
on the bivariate Poisson model. In addition to the paradigm change,
there are three major modifications on the parameterization. First
of all, the distribution for scores are truncated: scores greater
than four are treated as the same category. The authors argued that
the truncation reduces the extreme case where one team scores many
goals. Secondly, the final scores $S_{i}$ and $S_{j}$ are assumed
to be drawn from a mixture model:
\[
P(S_{i}=s_{i},S_{j}=s_{j})=(1-\epsilon)P_{DC}+\epsilon P_{Avg}
\]
The component $P_{DC}$ is the truncated version of the Dixon-Coles
model, and the component $P_{Avg}$ is a truncated bivariate Poisson
distribution (\ref{eq:DC_prob}) with $\mu$ and $\lambda$ equal
to the average value across all teams. Thus, the mixture model encourages
a reversion to the mean. Lastly, the attack parameters $\alpha$ and
defense parameters $\beta$ for each team changes over time following
a Brownian motion. The temporal dependence between match outcomes
are reflected by the change in parameters. This model does not have
an analytical posterior for parameters. The Bayesian inference procedure
is carried out via Markov Chain Monte Carlo method.

\citet{crowder2002dynamic} proposed another Bayesian formulation
of the bivariate Poisson model based on the Dixon-Coles model. The
parametric form remains unchanged, but the attack parameters $\alpha_{i}$'s
and defense parameter $\beta_{i}'s$ changes over time following an
AR(1) process. Again, the model does not have an analytical posterior.
The authors proposed a fast variational inference procedure to conduct
the inference.

\citet{baio2010bayesian} proposed a further extension to the bivariate
Poisson model proposed by~\citet{karlis2003analysis}. The authors
noted that the correlation between final scores are parametrized explicitly
in previous models, which seems unnecessary in the Bayesian setting.
In their proposed model, both scores are \textit{conditionally} independent
given an unobserved latent variable. This hierarchical structure naturally
encodes the \textit{marginal} dependence between the scores.

\subsection{Feature-based machine learning predictors\label{sub:Feature-based-machine-learning}}

In recent publications, researchers reported that machine learning
models achieved good prediction results for the outcomes of competitive
team sports. The strengths of the machine learning approach lie in
the model-agnostic and data-centric modelling using available off-shelf methodology,
as well as the ability to incorporate features in model building.

In this branch of research, the prediction problems are usually studied as a supervised classification problem,
either binary (home team win/lose or win/other), or ternary, i.e., where the outcome of a match falls into three distinct classes: home team
win, draw, and home team lose.

%A less-explored approach is to treat
%the match outcomes as ordinal variables: home team win $\succ$ draw
%$\succ$ away team win. This approach assumes that draw is a middle
%outcome, and it might be a reasonable assumption in many cases.

\citet{liu2010beating} applied logistic regression, support vector
machines with different kernels, and AdaBoost to predict NCAA football
outcomes. For this prediction problem, the researchers hand crafted
210 features.

\citet{hucaljuk2011predicting} explored more machine learning predictors
in the context of sports prediction. The predictors include naïve
Bayes classifiers, Bayes networks, LogitBoost, k-nearest neighbors, Random forest,
and artificial neural networks. The models are trained on 20 features
derived from previous match outcomes and 10 features designed subjectively
by experts (such as team's morale).

\citet{odachowski2012using} conducted a similar study. The predictors
are commercial implementations of various Decision Tree and ensembled
trees algorithms as well as a hand-crafted Bayes Network. The models
are trained on a subset of 320 features derived form the time series
of betting odds. In fact, this is the only study so far where the
predictors have no access to previous match outcomes.

\citet{kampakis2014using} explored the possibility of predicting
match outcome from Tweets. The authors applied naïve Bayes classifiers, Random
forests, logistic regression, and support vector machines to a feature
set composed of 12 match outcome features and a number of Tweets features.
The Tweets features are derived from unigrams and bigrams of the Tweets.

\subsection{Evaluation methods used in previous studies\label{sub:Evaluation-methods-used}}

In all studies mentioned in this section, the authors validated their
new model on a real data set and showed that the new model performs
better than an existing model. However, complication arises when we
would like to aggregate and compare the findings made in different
papers. Different studies may employ different validation settings,
different evaluation metrics, and different data sets. We report on this
with a focus on the following, methodologically crucial aspects:
\begin{enumerate}
\item[(i)] Studies may or may not include a well-chosen benchmark for comparison.
If this is not done, then it may not be concluded that the new method outperforms
the state-of-art, or a random guess.
\item[(ii)] Variable selection or hyper-parameter tuning procedures
may or may not be described explicitly. This may raise doubts about the validity
of conclusions, as ``hand-tuning'' parameters is implicit overfitting,
and may lead to underestimate the generalization error in validation.
\item[(iii)] Last but equally importantly, some studies do not
report the error measure on evaluation metrics (standard deviation
or confidence interval). In these studies, we cannot rule out the
possibility that the new model is outperforming the baselines just
by chance.
\end{enumerate}

In table~\ref{tab:Evaluation-methods-used}, we summarize the benchmark evaluation
methodology used in previous studies. One may remark that the size of testing
data sets vary considerably across different studies, and most studies
do not provide a quantitative assessment on the evaluation metric.
We also note that some studies perform the evaluation on the training
data (i.e., in-sample). Without further argument, these evaluation results
only show the goodness-of-fit of the model on the training data, as they do not provide
a reliable estimate of the expected predictive performance (on unseen data).

\begin{landscape}
\begin{table}[H]
\begin{centering}
\begin{tabular}{|>{\centering}p{4.5cm}|c|c|c|>{\centering}p{5.2cm}|c|c|c|c|}
\hline
{\small{}Study} & {\small{}Validation} & {\small{}Tuning} & {\small{}Task} & {\small{}Metrics} & {\small{}Baseline} & {\small{}Error} & {\small{}Train} & Test\tabularnewline
\hline
\hline
\citet{lasek2013predictive} & On-line & Yes & Binary & Brier score, Binomial divergence & Yes & Yes & NA & 979\tabularnewline
\hline
\citet{maher1982modelling} & In-sample & No & Scores & $\chi^{2}$ statistic & No & No & 5544 & NA\tabularnewline
\hline
\citet{dixon1997modelling} & No & No & Scores & Non-standard & No & No & NA & NA\tabularnewline
\hline
\citet{karlis2003analysis} & In-sample & Bayes & Scores & AIC, BIC & No & No & 615 & NA\tabularnewline
\hline
\citet{goddard2005regression} & Custom & Bayes & Scores & log-loss & No & No & 6930 & 4200\tabularnewline
\hline
\citet{rue2000prediction} & Custom & Bayes & Scores & log-loss & Yes & No & 280 & 280\tabularnewline
\hline
\citet{crowder2002dynamic} & On-line & Bayes & Tenary & Accuracy & No & No & 1680 & 1680\tabularnewline
\hline
\citet{baio2010bayesian} & Hold-out & Bayes & Scores & Not reported & No & No & 4590 & 306\tabularnewline
\hline
\citet{liu2010beating} & Hold-out & No & Binary & Accuracy & Yes & No & 480 & 240\tabularnewline
\hline
\citet{hucaljuk2011predicting} & Custom & Yes & Binary & Accuracy, F1 & Yes & No & 96 & 96\tabularnewline
\hline
\citet{odachowski2012using} & 10-fold CV & No & Tenary & Accuracy & Yes & No & 1116 & 1116\tabularnewline
\hline
\citet{kampakis2014using} & LOO-CV & No & Binary & Accuracy, Cohen’s kappa & No & Yes & NR & NR\tabularnewline
\hline
\end{tabular}
\par\end{centering}

\caption{Evaluation methods used in previous studies: the column \textquotedbl{}Validation\textquotedbl{}
lists the validation settings (\textquotedbl{}Hold-out\textquotedbl{}
uses a hold out test set, \textquotedbl{}10-fold CV\textquotedbl{} refers
means 10-fold cross validation, \textquotedbl{}LOO-CV\textquotedbl{}
means leave-one-out cross validation, \textquotedbl{}On-line\textquotedbl{}
means that on-line prediction strategies are used and validation is with a rolling horizon, \textquotedbl{}In-sample\textquotedbl{}
means prediction is validated on the same data the model was computed on, \textquotedbl{}Custom\textquotedbl{}
refers to a customized evaluation method); the column \textquotedbl{}Tuning\textquotedbl{}
lists whether the hyper-parameter tuning method is reported. The Bayesian methods' parameters
are ``tuned'' by the usual Bayesian update; \textquotedbl{}Task\textquotedbl{}
specifies the prediction task, Binary/Ternary = Binary/Ternary classification, Scores = prediction of final scores;
the column \textquotedbl{}Metric\textquotedbl{}
lists the evaluation metric(s) reported; \textquotedbl{}Baseline\textquotedbl{}
specifies whether baselines are reported; \textquotedbl{}Error\textquotedbl{}
specifies whether estimated error of the evaluation metric is reported;
\textquotedbl{}Test\textquotedbl{} specifies the number of data points in the test set;
\textquotedbl{}Train\textquotedbl{} specifies the number of data points in the training set.
For both training and test set, ``NA'' means that the number does not apply in the chosen set-up, for example because there was no test set;
``NR'' means that it does apply and should have been reported but was not reported in the reference.\label{tab:Evaluation-methods-used}}

\end{table}
\end{landscape}

\newpage
\section{Extending the Bradley-Terry-\'{E}l\H{o} model\label{sec:Methods}}

In this section, we propose a new family of models for the outcome
of competitive team sports, the structured log-odds models. We will
show that both Bradley-Terry and \'{E}l\H{o} models belong to this family (section
\ref{sub:The-structured-log-odds}), as well as logistic regression.
We then propose several new models with added flexibility (section \ref{sub:Extensions-of-structured})
and introduce various training algorithms (section \ref{sub:Training-structured-log-odds}
and \ref{sub:Regularized-log-odds-matrix}).

\subsection{The structured log-odds model\label{sub:The-structured-log-odds}}

Recall our principal observations obtained from the joint discussion of Bradley-Terry and \'{E}l\H{o} models in Section~\ref{sub:The-probabilistic-interpretation}:
\begin{enumerate}
\item[(i)] The \'{E}l\H{o} system can be seen as a learning algorithm for a logistic odds model with latent variables, the Bradley-Terry model
(and hence, by extension, as a full fit/predict specification of a certain one-layer neural network).
\item[(ii)] The Bradley-Terry and \'{E}l\H{o} model may simultaneously be interpreted as Bernoulli observation models of a rank two matrix.
\item[(iii)] The gradient of the Bradley-Terry model's log-likelihood gives rise to a (novel) batch gradient and a single-outcome gradient ascent update.
A single iteration per-sample of the latter (with a fixed update constant) is \'{E}l\H{o}'s original update rule.
\end{enumerate}

We collate these observations in a mathematical model, and highlight relations to well-known model classes,
including the Bradley-Terry-\'{E}l\H{o} model, logistic regression, and neural networks.

\subsubsection{Statistical definition of structured log-odds models\label{sub:Motivation-and-definition}}

In the definition below, we separate added assumptions and notations for the general set-up, given in the paragraph
``Set-up and notation'', from model-specific assumptions, given in the paragraph ``model definition''.
Model-specific assumptions, as usual, need not hold for the ``true'' generative process, and the mismatch of the assumed model structure
to the true generative process may be (and should be) quantified in a benchmark experiment.

\paragraph{Set-up and notation.}

We keep the notation of Section~\ref{sec:Background-and-Related}; for the time being, we assume that there
is no dependence on time, i.e., the observations follow a generative joint random variable $(X_{ij},Y_{ij})$.
The variable $Y_{ij}$ models the outcomes of a pairing where home team $i$ plays against away team $j$.
We will further assume that the outcomes are binary home team win/lose = 1/0, i.e., $Y_{ij}\sim\Bern (p_{ij})$.
The variable $X_{ij}$ models features relevant to the pairing.
From it, we may single out features that pertain to a single team $i$, as a variable $X_i$.
Without loss of generality (for example, through introduction of indicator variables), we will assume that $X_{ij}$ takes values in $\RR^n$, and $X_i$ takes values in $\RR^m$.
We will write $X_{ij,1},X_{ij,2},\dots, X_{ij,n}$ and $X_{i,1},\dots, X_{i,m}$ for the components.

The two restrictive assumptions (independence of time, binary outcome) are temporary and are made for expository reasons.
We will discuss in subsequent sections how these assumptions may be removed.

We have noted that the double sub-index notation easily allows to consider $p_*$ in matrix form.
We will denote by $\bP$ to the (real) matrix with entry $p_{ij}$ in the $i$-th row and $j$-th column.
Similarly, we will denote by $\bY$ the matrix with entries $Y_{ij}$.
We do not fix a particular ordering of the entries in $\bP,\bY$ as the numbering of teams does not matter,
however the indexing needs to be consistent across $\bP,\bY$ and any matrix of this format that we may define later.

A crucial observation is that the entries of the matrix $\bP$ can be plausibly expected to not be arbitrary.
For example, if team $i$ is a strong team, we should expect
$p_{ij}$ to be larger for all $j$'s. We can make a similar argument
if we know team $i$ is a weak team. This means the entries in matrix
$\bP$ are not completely independent from each other (in an algebraic sense); in other words,
the matrix $\bP$ can be plausibly assumed to have an inherent structure.

Hence, prediction of $\bY$ should be more accurate if the correct structural assumption is made on $\bP$,
which will be one of the cornerstones of the structured log-odds models.

For mathematical convenience (and for reasons of scientific parsimony which we will discuss),
we will not directly endow the matrix $\bP$ with structure, but the matrix $\bL:= \logit (\bP),$
where as usual and as in the following, univariate functions are applied entry-wise
(e.g., $\bP = \sigma(\bL)$ is also a valid statement and equivalent to the above).

\paragraph{Model definition.}

We are now ready to introduce the structured log-odds models for competitive team sports.
As the name says, the main assumption of the model is that the log-odds matrix $L$ is a structured
matrix, alongside with the other assumptions of the Bradley-Terry-\'{E}l\H{o} model in Section~\ref{sub:The-probabilistic-interpretation}.

More explicitly, all assumptions of the structured log-odds model may be written as
\begin{eqnarray}
\bY & \sim & \text{Bernoulli}(\bP) \nonumber \\
\bP & = & \sigma(\bL) \label{eq:model_summary} \\
\bL &  & \mbox{satisfies certain structural assumptions} \nonumber
\end{eqnarray}

where we have not made the structural assumptions on $\bL$ explicit yet.
The matrix $\bL$ may depend on $X_{ij},X_i$, though a sensible model
may be already obtained from a constant matrix
$\bL$ with restricted structure. We will show that the Bradley-Terry and \'{E}l\H{o} models are of this subtype.

\paragraph{Structural assumptions for the log-odds.}
We list a few structural assumptions that may or may not be present in some form,
and will be key in understanding important cases of the structured log-odds models.
These may be applied to $\bL$ as a constant matrix to obtain the simplest class of log-odds models,
such as the Bradley-Terry-\'{E}l\H{o} model as we will explain in the subsequent section.\\

{\bf Low-rankness.} A common structural restriction for a matrix (and arguably the most scientifically or mathematically parsimonious one)
is the assumption of low rank: namely, that the rank of the matrix of relevance is
less than or equal to a specified value $r$. Typically, $r$ is far less than
either size of the matrix, which heavily restricts the number of (model/algebraic) degrees of freedom in an
$(m\times n)$ matrix from $mn$ to $r(m+n-r)$.
The low-rank assumption essentially reflects a belief that the unknown matrix is determined by only a small number
of factors, corresponding to a small number of prototypical rows/columns, with the small number being equal to $r$.
By the singular value decomposition theorem, any rank $r$ matrix $A\in \RR^{m\times n}$ may be written as
$$A = \sum_{k=1}^r \lambda_k\cdot u^{(k)}\cdot \left(v^{(k)}\right)^\top,\quad\mbox{or, equivalently,}\quad A_{ij} = \sum_{k=1}^r \lambda_k\cdot u^{(k)}_i \cdot v^{(k)}_j$$
for some $\lambda_k\in \RR$, pairwise orthogonal $u^{(k)}\in \RR^m$, pairwise orthogonal $v^{(k)}\in \RR^n$;
equivalently, in matrix notation, $A = U\cdot \Lambda \cdot V^\top$ where $\Lambda\in \RR^{r\times r}$ is diagonal, and $U^\top U = V^\top V = I$ (and where $U\in \RR^{m\times r}, V \in \RR^{n\times r}$, and $u^{(k)}, v^{(k)}$ are the rows of $U,V$).\\

{\bf Anti-symmetry.} A further structural assumption is symmetry or anti-symmetry of a matrix.
Anti-symmetry arises in competitive outcome prediction naturally as follows:
if all matches were played on neutral fields (or if home advantage is modelled separately),
one should expect that $p_{ij}=1-p_{ji}$, which means the probability
for team $i$ to beat team $j$ is the same regardless of where the
match is played (i.e., which one is the home team).
Hence,
$$\bL_{ij} = \logit p_{ij} = \log \frac{p_{ij}}{1-p_{ij}} = \log \frac{1-p_{ji}}{p_{ji}} = -\logit p_{ji} = -\bL_{ji},$$
that is, $\bL$ is an anti-symmetric matrix, i.e., $\bL = - \bL^\top$.\\

{\bf Anti-symmetry and low-rankness.} It is known that any real antisymmetric matrix always has even rank~\citep{eves1980elementary}.
That is, if a matrix is assumed to be low-rank and anti-symmetric simultaneously, it will have rank $0$ or $2$ or $4$ etc.
In particular, the simplest (non-trivial) anti-symmetric low-rank matrices have rank $2$.
One can also show that any real antisymmetric matrix $A\in\RR^{n\times n}$ with rank $2r'$
can be decomposed as
%(\ref{eq:anti_decomp}).
\begin{equation}
A=\sum_{k=1}^{r'} \lambda_k\cdot \left(u^{(k)}\cdot \left(v^{(k)}\right)^{\top}-v^{(k)}\cdot \left(u^{(k)}\right)^{\top}\right) ,
\quad\mbox{or, equivalently,}\quad A_{ij} = \sum_{k=1}^{r'} \lambda_k\cdot \left(u^{(k)}_i \cdot v^{(k)}_j-u^{(k)}_j \cdot v^{(k)}_i\right)\label{eq:anti_decomp}
\end{equation}
for some $\lambda_k\in \RR$, pairwise orthogonal $u^{(k)}\in \RR^m$, pairwise orthogonal $v^{(k)}\in \RR^n$;
equivalently, in matrix notation, $A = U\cdot \Lambda \cdot V^\top - V\cdot \Lambda \cdot U^\top$
where $\Lambda\in \RR^{r\times r}$ is diagonal, and $U^\top U = V^\top V = I$ (and where $U, V \in \RR^{n\times r}$, and $u^{(k)}, v^{(k)}$ are the rows of $U,V$).\\

{\bf Separation.} In the above, in general, the factors $u^{(k)},v^{(k)}$ give rise to interaction constants (namely: $u^{(k)}_i\cdot v^{(k)}_j$) that are specific to the pairing.
To obtain interaction constants that only depend on one of the teams, one may additionally assume that one of the factors is constant,
or a vector of ones (without loss of generality from the constant vector). Similarly, a matrix with constant entries corresponds to an effect independent of the pairing.

\paragraph{Learning/fitting of structured log-odds models} will be discussed in Section~\ref{sub:Training-structured-log-odds}.
after we have established a number of important sub-cases and the full formulation of the model.

In a brief preview summary, it will be shown that the log-likelihood function has in essence the same form for
all structured log-odds models. Namely, for any parameter $\theta$ on which $\bP$ or $\bL$ may depend,
it holds for the (one-outcome log-likelihood) that
$$\ell (\theta|Y_{ij}) = Y_{ij}\log (p_{ij}) + (1-Y_{ij})\log (1-p_{ij}) = Y_{ij} \bL_{ij} + \log(1-p_{ij}).$$
Similarly, for its derivative one obtains
$$\frac{\partial \ell (\theta|Y_{ij})}{\partial \theta} = \frac{Y_{ij}}{p_{ij}}\cdot \frac{\partial p_{ij}}{\partial \theta} - \frac{1-Y_{ij}}{1-p_{ij}}\cdot \frac{\partial p_{ij}}{\partial \theta},$$
where the partial derivatives on the right hand side will have a different form for different structural assumptions,
while the general form of the formula above is the same for any such assumption.

Section~\ref{sub:Training-structured-log-odds} will expand on this for the full model class.

\subsubsection{Important special cases \label{sec:specialcases}}

We highlight a few important special types of structured log-odds models that we have already seen, or that are prototypical
for our subsequent discussion:\\

{\bf The Bradley-Terry-model} and via identification the \'{E}l\H{o} system are obtained under the structural assumption
that $\bL$ is anti-symmetric and of rank 2 with one factor vector of ones.

Namely, recalling equation \ref{eq:elo_prob2}, we recognize that the log-odds
matrix $\bL$ in the Bradley-Terry model is
given by $\bL_{ij}=\theta_{i}-\theta_{j}$,
where $\theta_{i}$ and $\theta_{j}$ are the \'{E}l\H{o} ratings.
Using the rule of matrix multiplication, one can verify that this is equivalent to
$$
\bL=\theta\cdot\OOne^{\top}-\OOne\cdot\theta^{\top}
$$
where $\OOne$ is a vector of ones and $\theta$ is the vector
of \'{E}l\H{o} ratings. For general $\theta$, the
log-odds matrix will have rank two (general = except if $\theta_i=\theta_j$ for all $i,j$). \\

By the exposition above, making the three assumptions is equivalent to positing the Bradley-Terry or \'{E}l\H{o} model.
Two interesting observations may be made:
First, the ones-vector being a factor entails that the winning chance
depends only on the difference between the team-specific ratings $\theta_i,\theta_j$, without any further interaction term.
Second, the entry-wise exponential of $\bL$ is a matrix of rank (at most) one.\\

{\bf The popular \'{E}l\H{o} model with home advantage} is obtained from the Bradley-Terry-\'{E}l\H{o} model under the structural assumption
that $\bL$ is a sum of low-rank matrix and a constant; equivalently, from an assumption
of rank 3 which is further restricted by fixing some factors to each other or to vectors of ones.

More precisely, from equation~\ref{eq:Elo_gen}, one can recognize that for the
\'{E}l\H{o} model with home advantage, the log-odds matrix decomposes as
$$
\bL=\theta\cdot\OOne^{\top}-\OOne\cdot\theta^{\top}+h\cdot \OOne\cdot\OOne^{\top}
$$
Note that the log-odds matrix is no longer antisymmetric due to the constant term
with home advantage parameter $h$ that is (algebraically) independent of the playing teams.
Also note that the anti-symmetric part, i.e., $\frac{1}{2}(\bL + \bL^\top)$,
is equivalent to the constant-free \'{E}l\H{o} model's log-odds, while the symmetric
part, i.e., $\frac{1}{2}(\bL - \bL^\top),$ is exactly the new constant home advantage term.\\

{\bf More factors: full two-factor Bradley-Terry-\'{E}l\H{o} models} may be obtained by dropping the
separation assumption from either Bradley-Terry-\'{E}l\H{o} model, i.e.,
keeping the assumption of anti-symmetric rank two, but allowing
an arbitrary second factor not necessarily being the vector of ones.
The team's competitive strength is then determined by two interacting factors
$u$, $v$, as
\begin{equation}
\bL =u\cdot v^{\top}-v\cdot u^{\top}\label{eq:fac2_log_odds}.
\end{equation}
Intuitively, this may cover, for example, a situation where the benefit from being much better may be
smaller (or larger) than being a little better, akin to a discounting of extremes.
If the full two-factor model predicts better than the Bradley-Terry-\'{E}l\H{o} model, it may certify for
different interaction in different ranges of the \'{E}l\H{o} scores.
A home advantage factor (a constant) may or may not be added, yielding a model of total rank 3.\\

{\bf Raising the rank: higher-rank Bradley-Terry-\'{E}l\H{o} models} may be obtained by
model by relaxing assumption of rank 2 (or 3) to higher rank.
We will consider the next more expressive model, of rank four.
The \emph{rank four Bradley-Terry-\'{E}l\H{o} model} which we will consider will add
a full anti-symmetric rank two summand to the log-odds matrix, which
hence is assumed to have the following structure:
\begin{equation}
\bL=u\cdot v^{\top}-v\cdot u^{\top}+\theta\cdot\OOne^{\top}-\OOne\cdot\theta^{\top}\label{eq:rank4_log_odds}
\end{equation}
The team's competitive strength is captured by
three factors $u$, $v$ and $\theta$; note that we have kept the vector of ones as a factor.
Also note that setting either of $u,v$ to $\OOne$ would \emph{not} result in a model extension
as the resulting matrix would still have rank two.
The rank-four model may intuitively make sense if there are (at least) two distinguishable qualities
determining the outcome - for example physical fitness of the team and strategic competence.
Whether there is evidence for the existence of more than one factor, as opposed to assuming
just a single one (as a single summary quantifier for good vs bad) may be checked by comparing predictive capabilities of the respective models.
Again, a home advantage factor may be added, yielding a log-odds matrix of total rank 5.

We would like to note that a mathematically equivalent model, as well as models with more factors, have already been considered by~\citet{stanescu2011rating},
though without making explicit the connection to matrices which are of low rank, anti-symmetric or structured in any other way.\\

{\bf Logistic regression} may also be obtained as a special case of
structured log-odds models. In the simplest form of logistic regression,
the log-odds matrix is a linear functional in the features.
Recall that in the case of competitive outcome prediction, we consider
pairing features $X_{ij}$ taking values in $\RR^n$, and team features $X_i$ taking values in $\RR^m$.
We may model the log-odds matrix as a linear functional in these, i.e., model that
$$
\bL_{ij} = \langle \lambda^{(ij)}, X_{ij}\rangle + \langle \beta^{(i)}, X_{i}\rangle + \langle \gamma^{(j)}, X_{j}\rangle + \alpha,
$$
where $\lambda^{(ij)}\in \RR^n, \beta^{(i)},\gamma^{(j)}\in \RR^m, \alpha\in \RR$.
If $\lambda^{(ij)} = 0$, we obtain a simple two-factor logistic regression model.
In the case that there is only two teams playing only with each other, or (the mathematical correlate of) a single team playing only with itself,
the standard logistic regression model is recovered.

Conversely, a way to obtain the Bradley-Terry model as a special case of
classical logistic regression is as follows:
consider the indicator feature $X_{ij}:= e_i - e_j$.
With a coefficient vector $\beta$, the logistic odds will be
$\bL_{ij}=\langle \beta, X_{ij}\rangle = \beta_{i}-\beta_{j}$.
In this case, the coefficient vector
$\beta$ corresponds to a vector of \'{E}l\H{o} ratings.

Note that in the above formulation, the coefficient vectors $\lambda^{(ij)}, \beta^{(i)}$ are explicitly allowed to depend on the teams.
If we further allow $\alpha$ to depend on both teams, the model includes the Bradley-Terry-\'{E}l\H{o} models above as well; we could also
make the $\beta$ depend on both teams.
However, allowing the coefficients to vary in full generality is not very sensible, and as for the constant term which
may yield the \'{E}l\H{o} model under specific structural assumptions, we need to endow all model parameters with
structural assumptions to prevent combinatorial explosion of parameters and overfitting.

These subtleties in incorporating features, and more generally
how to combine features with hidden factors
will be discussed in the separate, subsequent Section~\ref{sub:covariate}.

\subsubsection{Connection to existing model classes}

Close connections to three important classes of models become apparent through the discussion in the previous sections:\\

{\bf Generalized Linear Models} generalize both linear and log-linear models (such as the Bradley-Terry model) through
so-called link functions, or more generally (and less classically) link distributions,
combined with flexible structural assumptions on the target variable.
The generalization aims at extending prediction with linear functionals through
the choice of link which is most suitable for the target~\citep[for an overview, see][]{mccullagh1989generalized}.

Particularly relevant for us are generalized linear models for ordinal outcomes which includes the
ternary (win/draw/lose) case, as well as link distributions for scores. Some existing extensions of this type,
such as the ternay outcome model of~\cite{rao1967ties} and the score model of~\citep{maher1982modelling},
may be interpreted as specific choices of suitable linking distributions.
How these ideas may be used as a component of structured log-odds models will be discussed in Section~\ref{sub:Extensions-of-structured}.\\

{\bf Neural Networks} (vulgo ``deep learning'') may be seen as a generalization of logistic regression which is
mathematically equivalent to a single-layer network with softmax activation function. The generalization is achieved
through functional nesting which allows for non-linear prediction functionals, and greatly expands the capability of regression models to handle
non-linear features-target-relations \citep[for an overview, see][]{schmidhuber2015deep}.

A family of ideas which immediately transfers to our setting are strategies for training and model fitting.
In particular, on-line update strategies as well as training in batches and epochs yields a natural
and principled way to learn Bradley-Terry-\'{E}l\H{o} and log-odds models in an on-line setting or to potentially improve its predictive power in a
static supervised learning setting.
A selection of such training strategies for structured log-odds models will be explored in Section~\ref{sub:Training-structured-log-odds}.
This will not include variants of stochastic gradient descent which we leave to future investigations.

It is also beyond the scope of this manuscript to explore the implications of using multiple layers
in a competitive outcome setting, though it seems to be a natural idea given the closeness of the model classes
which certainly might be worth exploring in further research.\\

{\bf Low-rank Matrix Completion} is the supervised task of filling in some
missing entries of a low-rank matrix, given others and the information that the rank is small.
Many machine learning applications can be viewed as estimation
or completion of a low-rank matrix, and different solution strategies exist~\citep{CanRec09,CandesTao,NegWai11,KMO10,Mek09,so2007theory,vounou2010discovering,kiraly2015algebraic}.

The feature-free variant of structured log-odds models (see Section~\ref{sub:Motivation-and-definition}) may be regarded as a
low-rank matrix completion problem: from observations of $Y_{ij}\sim\Bern(\sigma(\bL_{ij})),$ for $(i,j)\in E$ where the set of observed pairings
$E$ may be considered as the set of observed positions, estimate the underlying low-rank matrix $\bL$, or
predict $Y_{k\ell}$ for some $(k,\ell)$ which is possibly not contained in $E$.

One popular low-rank matrix completion strategy in estimating model parameters or completing missing entries uses the idea of replacing the discrete rank constraint
by a continuous spectral surrogate constraint, penalizing not rank but the nuclear norm ( = trace norm = 1-Schatten-norm)
of the matrix modelled to have low rank~\citep[an early occurrence of this idea may be found in][]{SreShr05}.
The advantage of this strategy is that no particular rank needs to be a-priori assumed, instead the objective implicitly selects a low rank
through a trade-off with model fit. This strategy will be explored in Section~\ref{sub:Regularized-log-odds-matrix} for the structured log-odds models.

Further, identifiability of the structured log-odds models is closely linked to the question whether a given entry of a low-rank matrix
may be reconstructed from those which have been observed.
Somewhat straightforwardly, one may see that reconstructability in the algebraic sense, see~\citep{kiraly2015algebraic},
is a necessary condition for identifiability under respective structure assumptions.
However, even though many results of~\citep{kiraly2015algebraic} directly generalize,
completability of anti-symmetric low-rank matrices with or without vectors of ones being factors has not been studied explicitly in literature to our knowledge,
hence we only point this out as an interesting avenue for future research.

We would like to note that a more qualitative and implicit mention of this, in the form of noticing connection to the general area of collaborative filtering,
is already made in~\cite[Section~6.3 of][]{paterek2012predicting}, in reference to the multi-factor models studied by~\citet{stanescu2011rating}.

\subsection{Predicting non-binary labels with structured log-odds models\label{sub:Extensions-of-structured}}

In Section~\ref{sub:The-structured-log-odds}, we have not introduced all
aspects of structured log-odds models in favour of a clearer exposition.
In this section, we discuss these aspects
that are useful for the domain application more precisely, namely:

\begin{enumerate}
\item[(i)] How to use features in the prediction.
\item[(ii)] How to model ternary match outcomes (win/draw/lose) or score outcomes.
\item[(iii)] How to train the model in an on-line setting with a batch/epoch strategy.
\end{enumerate}

For point (i) ``using features'', we will draw from the structured log-odds models' closeness to logistic regression;
the approach to (ii) ``general outcomes'' may be treated by choosing an appropriate link function as with generalized linear models;
for (iii), parallels may be drawn to training strategies for neural networks.

\subsubsection{The structured log-odds model with features\label{sub:covariate}}

As highlighted in Section~\ref{sec:specialcases}, pairing features $X_{ij}$ taking values in $\RR^n$, and team features $X_i$ taking values in $\RR^m$
may be incorporated by modelling the log-odds matrix as
\begin{equation}
\bL_{ij} = \langle \lambda^{(ij)}, X_{ij}\rangle + \langle \beta^{(i)}, X_{i}\rangle + \langle \gamma^{(j)}, X_{j}\rangle + \alpha_{ij}, \label{eq:logft-oneentry}
\end{equation}
where $\lambda^{(ij)}\in \RR^n, \beta^{(i)},\gamma^{(j)}\in \RR^m, \alpha_{ij}\in \RR$. Note that differently from the simpler exposition in
Section~\ref{sec:specialcases}, we allow all coefficients, including $\alpha_{ij}$, to vary with $i$ and $j$.

Though, allowing $\lambda^{(ij)}$ and $\beta^{(i)},\gamma^{(j)}$ to vary completely freely may lead to over-parameterisation or overfitting,
similarly to an unrestricted (full rank) log-odds matrix of $\alpha_{ij}$ in the low-rank \'{E}l\H{o} model,
especially if the number of distinct observed pairings is of similar magnitude as the number of total observed outcomes.

Hence, structural restriction of the degrees of freedom may be as important for the feature coefficients as for the constant term.
The simplest such assumption is that all $\lambda^{(ij)}$ are equal, all $\beta^{(i)}$ are equal, and all $\gamma^{(j)}$ are equal, i.e., assuming that
$$
\bL_{ij} = \langle \lambda, X_{ij}\rangle + \langle \beta, X_{i}\rangle + \langle \gamma, X_{j}\rangle + \alpha_{ij},
$$
for some $\lambda\in \RR^n, \beta,\gamma\in \RR^m,$ and where $\alpha_{ij}$ may follow the assumptions of the feature-free log-odds models.
This will be the main variant which will refer to as the structured log-odds model with features.

However, the assumption that constants are independent of the pairing $i,j$ may be too restrictive, as it may be plausible
that, for example, teams of different strength profit differently from or are impaired differently by the same circumstance,
e.g., injury of a key player.

To address such a situation, it is helpful to re-write Equation~\ref{eq:logft-oneentry} in matrix form:
$$
\bL = \blambda \circ_3 \bX + \bbeta \cdot \bX_*^\top + \bX_*\cdot \bgamma^\top + \balpha,
$$
where $\bX_*$ is the matrix whose rows are the $X_i$, where $\bbeta$ and $\bgamma$ are matrices whose rows are the $\beta^{(i)},\gamma^{(j)}$, and where
$\balpha$ is the matrix with entries $\alpha_{ij}$.
The symbols $\blambda$ and $\bX$ denote tensors of degree 3 (= 3D-arrays)
whose $(i,j,k)$-th elements are $\lambda^{(ij)}_k$ and $X_{ij,k}$. The symbol $\circ_3$ stands for the index-wise product of degree-3-tensors which eliminates
the third index and yields a matrix, i.e.,
$$\left(\blambda \circ_3 \bX\right)_{ij} = \sum_{k=1}^n \lambda^{(ij)}_k\cdot X_{ij,k}.$$

A natural parsimony assumption for $\bgamma,\bbeta,\balpha$, and $\blambda$ is, again, that of low-rank.
For the matrices, $\bgamma,\bbeta,\balpha$, one can explore the same structural assumptions as in Section~\ref{sub:Motivation-and-definition}:
low-rankness and factors of one are reasonable to assume for all three, while anti-symmetry seems natural for $\balpha$ but not for $\bbeta,\bgamma$.

A low tensor rank (Tucker or Waring) appears to be a reasonable assumption for $\blambda$. As an ad-hoc definition of tensor (decomposition) rank of $\blambda$,
one may take the minimal $r$ such that there is a decomposition into real vectors $u^{(i)},v^{(i)},w^{(i)}$ such that
$$\blambda_{ijk} = \sum_{\ell=1}^r u^{(\ell)}_i\cdot v^{(\ell)}_j\cdot w^{(\ell)}_k.$$
Further reasonable assumptions are anti-symmetry in the first two indices, i.e., $\blambda_{ijk} = - \blambda_{jik}$, as well as some factors $u^{(\ell)}, v^{(\ell)}$ being vectors of ones.

Exploring these possible structural assumptions on the coefficients of features in experiments is possibly interesting both from
a theoretical and practical perspective, but beyond the scope of this manuscript.
Instead, we will restrict ourselves to the case of $\blambda = 0$, of $\bbeta$ and $\bgamma$ having the same entry each, and $\balpha$
following one of the low-rank assumptions in structural assumptions as in Section~\ref{sub:Motivation-and-definition} as in the feature-free model.

We would like to note that variants of the Bradley-Terry model with features have already been proposed and implemented in the \texttt{BradleyTerry2} package for R~\cite{firth2012bradley}, though isolated from other aspects of the Bradley-Terry-\'{E}l\H{o} model class such as modelling draws,
or structural restrictions on hidden variables or the coefficient matrices and tensors, and the \'{E}l\H{o} on-line update.

\subsubsection{Predicting ternary outcomes\label{sub:Modeling-tenary-outcomes}}

This section addresses the issue of modeling draws raised in \ref{sub:Limitations-Elo}.
When it is necessary to model draws, we assume that the outcome of
a match is an ordinal random variable of three so-called levels: win $\succ$
draw $\succ$ lose. The draw is treated as a middle outcome. The extension
of structured log-odds model is inspired by an extension of logistic
regression: the Proportional Odds model.

The Proportional Odds model is a well-known family of models for ordinal
random variables~\citep{mccullagh1980regression}. It extends the
logistic regression to model ordinary target variables. The model
parameterizes the logit transformation of the cumulative probability
as a linear function of features. The coefficients associated
with feature variables are shared across all levels, but there is
an intercept term $\alpha_{k}$ which is specific to a certain level.
For a generic feature-label distribution $(X,Y)$, where $X$ takes values in $\RR^n$
and $Y$ takes values in a discrete set $\calY$ of ordered levels, the proportional odds model
may be written as
\[
\log\left(\frac{P(Y \succ k)}{P(Y \preceq k)}\right)=\alpha_{k}+\langle \beta, X\rangle
\]
where $\beta\in\RR^n, \alpha_k\in \RR$, and $k\in \calY$.
The model is called Proportional Odds model because the odds for any
two different levels $k$, $k'$, given an observed feature set, are proportional with a constant
that does not depend on features; mathematically,
\[
\left(\frac{P(Y\succ k)}{P(Y\preceq k)}\right)/\left(\frac{P(Y\succ k')}{P(Y\preceq k')}\right)=\exp(\alpha_{k}-\alpha_{k'})
\]

Using a similar formulation in which we closely follow~\citet{rao1967ties}, the structured log-odds model can be
extended to model draws, namely by setting
\begin{eqnarray*}
\log\left(\frac{P(Y_{ij}=\text{win})}{P(Y_{ij}=\text{draw})+P(Y_{ij}=\text{lose})}\right) & = & \bL_{ij}\\
\log\left(\frac{P(Y_{ij}=\text{draw})+P(Y_{ij}=\text{win})}{P(Y_{ij}=\text{lose})}\right) & = & \bL_{ij}+\phi
\end{eqnarray*}
where $\bL_{ij}$ is the entry in structured log-odds matrix and $\phi$
is a free parameter that affects the estimated probability of a draw.
Under this formulation, the probabilities for different outcomes are
given by
\begin{eqnarray*}
P(Y_{ij}=\text{win}) & = & \sigma(\bL_{ij})\\
P(Y_{ij}=\text{lose}) & = & \sigma(-\bL_{ij}-\phi)\\
P(Y_{ij}=\text{draw}) & = & \sigma(-\bL_{ij})-\sigma(-\bL_{ij}-\phi)
\end{eqnarray*}

Note that this may be seen as a choice of ordinal link distribution in a ``generalized'' structured odds model,
and may be readily combined with feature terms as in Section~\ref{sub:covariate}.

\subsubsection{Predicting score outcomes\label{sub:Using-score-difference}}

Several models have been considered in Section \ref{sub:Limitations-Elo}
that use score differences to update the \'{E}l\H{o} ratings.
In this section, we derive a principled way to predict scores, score differences
and/or learn from scores or score differences.

Following the analogy to generalized linear models, we will be able to
tackle this by using a suitable linking distribution, the model can utilize additional
information in final scores.

The simplest natural assumption one may make on scores is obtained from assuming
a dependent scoring process, i.e., both home and away team's scores
are Poisson-distributed with a team-dependent parameter and possible correlation.
This assumption is frequently made in literature~\citep{maher1982modelling,dixon1997modelling,crowder2002dynamic}
and eventually leads to a (double) Poisson regression when combined with structured log-odds models.

The natural linking distributions for differences of scores
are Skellam distributions which are obtained as difference distributions of two (possibly correlated) Poisson
distributions~\citep{skellam1945frequency}, as it has been suggested by~\citet{karlis2009bayesian}.

In the following, we discuss only the case of score differences in detail,
predicting both team's score distributions can be obtained similarly
as predicting the correlated Poisson variables with the respective parameters instead of the Skellam difference distribution.

We first introduce some notation.
As a difference of Poisson distributions whose support is $\NN$, the support of a Skellam distribution is the set of integers $\ZZ$.
The probability
mass function of Skellam distributions takes two positive parameters $\mu_{1}$ and $\mu_{2}$,
and is given by
\[
P(z|\mu_{1},\mu_{2})=e^{-(\mu_{1}+\mu_{2})}\left(\frac{\mu_{1}}{\mu_{2}}\right)^{z/2}I_{|z|}(2\sqrt{\mu_{1}\mu_{2}})
\]
where $I_{\alpha}$ is the modified Bessel function of first kind with parameter $\alpha$, given
by
\[
I_{\alpha}(x):=\sum_{k=0}^{\infty}\frac{1}{k!\cdot \Gamma(\alpha+k+1)}\cdot \left(\frac{x}{2}\right)^{2k + \alpha}
\]
If random variables $Z_{1}$ and $Z_{2}$ follow Poisson distributions with mean parameters $\lambda_{1}$ and $\lambda_{2}$ respectively,
and their correlation is $\rho=\Corr (Z_1,Z_2)$, then their difference $\tilde{Z}=Z_{1}-Z_{2}$
follows a Skellam distribution with mean parameters $\mu_{1}=\lambda_{1}-\rho\sqrt{\lambda_{1}\lambda_{2}}$
and $\mu_{2}=\lambda_{2}-\rho\sqrt{\lambda_{1}\lambda_{2}}$.

Now we are ready to extend the structured log-odds model to incorporate
historical final scores. We will use a Skellam distribution as the
linking distribution: we assume that the score difference of a match
between team $i$ and team $j$, that is, $Y_{ij}$ (taking values in $\calY = \ZZ$),
follows a Skellam distribution with (unknown) parameter
$\exp(\bL_{ij})$ and $\exp(\bL'_{ij})$.

Note that hence there are now \emph{two} structured $\bL,\bL'$ , each of which
may be subject to constraints such as in Section~\ref{sub:Motivation-and-definition},
or constraints connecting them to each other, and each of which may
depend on features as outlined in Section~\ref{sub:covariate}.

A simple (and arguably the simplest sensible) structural assumption is that $\bL^\top= \bL'$,
is rank two, with factors of ones, as follows:
$$\bL = \OOne\cdot u^\top + v\cdot \OOne^\top;$$
equivalently, that $\exp(\bL)$ has rank one and only non-negative entries.

As mentioned above, features such as home
advantage may be added to the structured parameter matrix $\bL$ or $\bL'$ using the
way introduced in Section~\ref{sub:covariate}.

Also note that the above yields a strategy to make ternary predictions while training on the scores.
Namely, a prediction for ternary match outcomes may simply be derived from predicted
score differences $\tilde{Y}_{ij}$, through defining
\begin{eqnarray*}
P(\text{win}) & = & P({Y}_{ij}>0)\\
P(\text{draw}) & = & P({Y}_{ij}=0)\\
P(\text{lose}) & = & P({Y}_{ij}<0)
\end{eqnarray*}
In contrast to the direct method in
Section~\ref{sub:Modeling-tenary-outcomes},
the probability of draw can now be calculated without introducing
an additional cut-off parameter.

\subsection{Training of structured log-odds models\label{sub:Training-structured-log-odds}}

In this section, we introduce batch and on-line learning strategies for structured log-odds models,
based on gradient descent on the parametric likelihood.

The methods are generic in the sense that the exact structural assumptions of the model will affect the
exact form of the log-likelihood, but not the main algorithmic steps.

\subsubsection{The likelihood of structured log-odds models}

We derive a number of re-occurring formulae for the likelihood of structured log-odds models.
For this, we will subsume all structural assumptions on $\bL$ in the form of a parameter $\theta$
on which $\bL$ may depend, say in the cases mentioned in Section~\ref{sec:specialcases}.
In each case, we consider $\theta$ to be a real vector of suitable length.

The form of the learning step(s) is slightly different depending on the chosen link function/distribution, hence
we start with our derivations in the case of binary prediction, where $\calY = \{1,0\}$, and discuss ternary
and score outcomes further below.\\

In the case of {\bf binary prediction}, it holds for the (one-outcome log-likelihood) that
\begin{align*}
\ell (\theta|X_{ij},Y_{ij})& = Y_{ij}\log (p_{ij}) + (1-Y_{ij})\log (1-p_{ij})\\
& = Y_{ij} \bL_{ij} + \log(1-p_{ij}) = Y_{ij} \bL_{ij} - \bL_{ij} + \log(p_{ij}).
\end{align*}
Similarly, for its derivative one obtains
\begin{eqnarray}
\frac{\partial \ell (\theta|X_{ij},Y_{ij})}{\partial \theta} & = & \frac{\partial}{\partial\theta} \left[Y_{ij}\log p_{ij}+\left(1-Y_{ij}\right)\log(1-p_{ij})\right] \nonumber \\
 & = & \left[\frac{Y_{ij}}{p_{ij}} - \frac{1-Y_{ij}}{1-p_{ij}}\right]\cdot \frac{\partial p_{ij}}{\partial \theta} \label{eq:derivative}\\
 & = & \left[Y_{ij}-p_{ij}\right]\cdot\frac{\partial}{\partial\theta}\bL_{ij} \nonumber
\end{eqnarray}
where we have used definitions for the first equality, the chain rule for the second, and for the last equality that
$$\frac{\partial }{\partial x} \sigma (x) = \sigma(x) (1-\sigma(x)),\;\mbox{hence}\;\; \frac{\partial }{\partial x} p_{ij} = p_{ij}(1-p_{ij})\frac{\partial }{\partial x} \bL_{ij}.$$
In all the above, derivatives with respect to $\theta$ are to be interpreted as (entry-wise) vector derivatives; equivalently, the equations hold for any
coordinate of $\theta$ in place of $\theta$.

As an important consequence of the above, the derivative of the log-likelihood
almost has the same form (\ref{eq:derivative}) for different model variants, and
differences only occur in the gradient
term $\frac{\partial}{\partial\theta_{i}}L_{ij}$; the term
$\left[Y_{ij}-p_{ij}\right]$ may be interpreted as a prediction residual, with $p_{ij}$ depending
 on $X_{ij}$ for a model with features. This fact enables
us to obtain unified training strategies for a variety of structured log-odds models.\\

For {\bf multiple class prediction} as in the ordinal or score case, the above
generalizes relatively straightforwardly. The one-outcome log-likelihood is given as
\begin{align*}
\ell (\theta|X_{ij},Y_{ij})& = \sum_{y\in \calY} Y_{ij}[y] \log p_{ij}[y]
\end{align*}
where, abbreviatingly, $p_{ij}[y] = P(Y_{ij} = y)$, and $Y_{ij}[y]$ is one iff $Y_{ij}$ takes the value $y$, otherwise zero.
For the derivative of the log-likelihood, one hence obtains
\begin{eqnarray}
\frac{\partial \ell (\theta|X_{ij},Y_{ij})}{\partial \theta} & = & \frac{\partial}{\partial\theta} \sum_{y\in \calY} Y_{ij}[y] \log (p_{ij}[y]) \nonumber \\
 & = & \sum_{y\in \calY} \frac{Y_{ij}[y]}{p_{ij}[y]}\cdot \frac{\partial p_{ij}[y]}{\partial \theta} \nonumber\\
 & = & \sum_{y\in \calY}\left[Y_{ij}[y]\cdot (1-p_{ij}[y])\right]\cdot\frac{\partial}{\partial\theta}\bL_{ij}[y], \nonumber
\end{eqnarray}
where $\bL_{ij}[y]:= \logit p_{ij}[y]$.

This is in complete analogy to the binary case, except for the very final cancellation which does not occur.
If $Y_{ij}$ is additionally assumed to follow a concrete distributional form (say Poisson or Skellam), the expression may be further simplified.
In the subsequent sections, however, we will continue with the binary case only, due to the relatively straightforward analogy through the above.

In either case, we note the similarity with back-propagation in neural networks, where
the derivatives $\frac{\partial}{\partial \theta}\bL_{ij}[y]$ correspond to a ``previous layer''. Though we would like to note
that differently from the standard multilayer perceptron, additional structural constraints on this layer
are encoded through the structural assumptions in the structured log-odds model.
Exploring the benefit of such constraints in general neural network layers is beyond the scope of this manuscript,
but a possibly interesting avenue to explore.

\subsubsection{Batch training of structured log-odds models\label{sub:Training-structured-log-odds.batch}
\label{sub:Two-stage-training-method}}

We now consider the case where a batch of multiple training outcomes
$\calD = \left\{\left(X_{i_1j_1}^{(1)},Y_{i_1j_1}^{(1)}\right),\dots,\left(X_{i_1j_1}^{(1)},Y_{i_Nj_N}^{(N)}\right)\right\}$
have been are observed, and we would like to train the model parameters the log-likelihood, compare the discussion in Section~\ref{sub:The-probabilistic-interpretation}.

In this case, the batch log-likelihood of the parameters $\theta$ and its derivative take the form
\begin{eqnarray}
\ell (\theta|\calD) &= &\sum_{k=1}^N \ell \left(\theta \middle|\left(X_{i_kj_k}^{(k)},Y_{i_kj_k}^{(k)}\right)\right)\\\nonumber
&= &\sum_{k=1}^N \left[ Y_{ij}^{(k)}\log \left(p_{ij}^{(k)}\right) + \left(1-Y_{ij}^{(k)}\right)\log \left(1-p_{ij}^{(k)}\right)\right]\\\nonumber
\frac{\partial}{\partial\theta}\ell (\theta|\calD) &= &\sum_{k=1}^N \left[Y_{i_kj_k}^{(k)}-p_{i_kj_k}^{(k)}\right]\cdot\frac{\partial}{\partial\theta}\bL_{i_kj_k}^{(k)}\label{eqn:batch_update}
\end{eqnarray}
Note that in general, both $p_{ij}^{(k)}$ and $\bL_{ij}^{(k)}$ will depend on the respective features $X_{i_kj_k}^{(k)}$ and the
parameters $\theta$, which is not made explicit for notational convenience.
The term $\left[Y_{i_kj_k}^{(k)}-p_{i_kj_k}^{(k)}\right]$ may again be interpreted as a sample of prediction residuals, similar to the one-sample case.

By the maximum likelihood method, the maximizer $\widehat{\theta} := \argmax_{\theta} \; \ell (\theta|\calD)$ is an estimate for the generative $\theta$.
In general, unfortunately, an analytic solution will not exist; nor will the optimization be convex, not even for the Bradley-Terry-\'{E}l\H{o} model.
Hence, gradient ascent and/or non-linear optimization techniques need to be employed.

An interesting property of the batch optimization is that a-priori setting a ``K-factor'' is not necessary.
While it may re-enter as the learning rate in a gradient ascent strategy, such parameters may be tuned
in re-sampling schemes such as k-fold cross-validation.

It also removes the need for a heuristic that determines new players' ratings (or more generally: factors),
as the batch training procedure may simply be repeated with such players' outcomes included.

\subsubsection{On-line training of structured log-odds models}\label{sub:Training-structured-log-odds.online}

In practice, the training data accumulate through time, so we need
to re-train the model periodically in order to capture new information.
That is, we would like to address the situation where training data $X_{ij}(t),Y_{ij}(t)$ are observed
at subsequent different time points.

The above-mentioned vicinity of structured log-odds models to neural networks
and standard stochastic gradient descent strategies directly yields a
family of possible batch/epoch on-line strategies for structured log-odds models.

To be more mathematically precise (and noting that the meaning of batch and epoch is not consistent across literature):
Let $\calD=\left\{\left(X^{(1)}_{i_1j_1}(t_1),Y^{(1)}_{i_1j_1}(t_1)\right),\dots, \left(X^{(N)}_{i_Nj_N}(t_N),Y^{(N)}_{i_Nj_N}(t_N)\right)\right\}$
be the observed training data points, at the (not necessarily distinct) time points $\calT = \{t_1,\dots, t_N\}$ (hence $\mathcal{T}$ can be a multi-set).

We will divide the time points into blocks $\calT_0,\dots, \calT_B$ in a sequential way,
i.e., such that $\cup_{i=0}^B \calT_i = \calT$, and for any two distinct $k,\ell$, either $x<y$ for all $x\in\mathcal{T}_k,y\in\calT_\ell$, or $x>y$
for all $x\in\mathcal{T}_k,y\in\calT_\ell$. These time blocks give rise to the training data
\emph{batches} $\calD_i:=\{(x,y)\in \calD\;:\; (x,y)\;\mbox{is observed at a time}\; t\in\calT_i\}$.
The cardinality of $\calD_i$ is called the \emph{batch size} of the $i$-th batch.
We single out the $0$-th batch as the ``initial batch''.

The stochastic gradient descent update will be carried out, for the $i$-th batch, $\tau_i$ times.
The $i$-th \emph{epoch} is the collection of all such updates using batch $\calD_i$, and $\tau_i$ is called the \emph{epoch size} (of epoch $i$).
Usually, all batches except the initial batch will have equal batch sizes and epoch sizes.

The general algorithm for the parameter update is summarized as stylized pseudo-code as Algorithm~\ref{alg:batch_epoch_training}.

\begin{algorithm}
\begin{algorithmic}[0]
\Require{learning rate $\gamma$}
\State Randomly initialize parameters $\theta$
\For{$i = 0: B$}
\State Read $\calD_i$
\For{$j = 1: \tau_i$}
\State Compute $\Delta:= \frac{\partial}{\partial\theta}\ell (\theta|\calD_i)$ as in Equation~\ref{eqn:batch_update}
\State $\theta \leftarrow \theta - \gamma\cdot \Delta$
\EndFor
\State Write/output $\theta$, e.g., for prediction or forecasting
\EndFor
\end{algorithmic}
\caption{Batch/epoch type on-line training for structured log-odds models\label{alg:batch_epoch_training}}
\end{algorithm}

Of course, any more sophisticated variant of stochastic gradient descent/ascent may be used here as well,
though we did not explore such possibilities in our empirical experiments and leave this for interesting future investigations.
Important such variants include re-initialization strategies, selecting the epoch size $\tau_i$ data-dependently by convergence criteria,
or employing smarter gradient updates, such as with data-dependent learning rates.

Note that the update rule applies for any structured log-odds model as long as
$\frac{\partial}{\partial\theta}\ell (\theta|\calD_i)$ is easily obtainable,
which should be the case for any reasonable parametric form and constraints.

Note that the online update rule may also be used to update, over time, structural model parameters such as home advantage and feature coefficients.
Of course, some parameters may also be regarded as classical hyper-parameters and tuned via grid or random search on a validation set.

There are multiple trade-offs involved in choosing the batches and epochs:

\begin{enumerate}
\item[(i)] Using more, possibly older outcomes vs emphasizing more recent outcomes.
        Choosing a larger epoch size will yield a parameter closer to the maximizer of the likelihood given the most recent batch(es).
        It is widely hypothesized that the team's performance changes gradually over time.
        If the factors change quickly, then more recent outcomes should be emphasized via larger epoch size.
        If they do not, then using more historical data via smaller epoch sizes is a better idea.
\item[(ii)] Expending less computation for a smooth update vs expending more computation for a more accurate update.
        Choosing a smaller learning rate will avoid ``overshooting'' local maximizers of the likelihood, or oscillations,
        though it will make a larger epoch size necessary for convergence.
\end{enumerate}

We single out multiple variants of the above to investigate the above trade-off and empirical merits of different on-line training strategies:

\begin{enumerate}
\item[(i)] {\bf Single-batch max-likelihood}, where there is only the initial batch ($B=0$), and a very large number of epochs (until convergence of the log-likelihood).
This strategy, in essence, disregards any temporal structure and is equivalent to the classical maximum likelihood approach under the given model assumptions.
It is the ``no time structure'' baseline, i.e., it should be improved upon for the claim that there is temporal structure.

\item[(ii)] {\bf Repeated re-training} is using re-training in regular intervals using the single-batch max-likelihood strategy.
Strictly speaking not a special case of Algorithm~\ref{alg:batch_epoch_training}, this is a less sophisticated and possibly much more computationally expensive baseline.

\item[(iii)] {\bf On-line learning} is Algorithm~\ref{alg:batch_epoch_training} with all batch and epoch sizes equal, parameters tuned on a validation set.
    This is a ``standard'' on-line learning strategy.

\item[(iv)] {\bf Two-stage training}, where the initial batch and epoch size is large, and all other batch and epoch sizes are equal, parameters tuned on a validation set.
    This is single-batch max-likelihood on a larger corpus of not completely recent historical data, with on-line updates starting only in the recent past.
    The idea is to get an accurate initial guess via the larger batch which is then continuously updated with smaller changes.
\end{enumerate}

In this manuscript, the most recent model will only be used to predict the labels/outcomes in the most recent batch.

\subsection{Rank regularized log-odds matrix estimation\label{sub:Regularized-log-odds-matrix}}

All the structured log-odds models we discussed so far made explicit
assumption about the structure of the log-odds matrix. An alternative
way is to encourage the log-odds matrix to be more structured by imposing
an implicit penalty on its complexity. In this way, there is no need to specify
the structure explicitly. The trade-off between the log-odds matrix's
complexity and its ability to explain observed data is tuned by validation
on evaluation data set.

The discussion will be based on the binary outcome model from Section~\ref{sub:The-structured-log-odds}.
Without any further
assumption about the structure of $\bL$ or $\bP$, the maximum
likelihood estimate for each $p_{ij}$ is given by
\[
\hat{p}_{ij}:=\frac{W_{ij}}{N_{ij}}
\]
where $W_{ij}$ is the number of matches in which team $i$ beats
team $j$, and $N_{ij}$ is the total number of matches between team
$i$ and team $j$. As we have assumed observations of wins/losses
to be independent, this immediately yields
$\hat{\bP} := \bW/\bN,$ as the maximum likelihood estimate for $\bP$,
where $\hat{\bP}, \bW,\bN,$ are the matrices with $\hat{p}_{ij},{W_{ij}},{N_{ij}}$
as entries and division is entry-wise.

Using the invariance of the maximum likelihood estimate under the bijective transformation
$\bL_{ij} = \logit (p_{ij})$, one obtains the maximum likelihood estimate for $\bL_{ij}$ as
\[
\hat{\bL}_{ij}=\log\left(\frac{\hat{p}_{ij}}{1-\hat{p}_{ij}}\right)= \log W_{ij} - \log W_{ji},
\]
or, more concisely, $\hat{\bL} = \log \bW - \log \bW^\top$, where the $\log$ is entry-wise.

We will call the matrix $\hat{\bL}$ the empirical log-odds matrix. It is worth noticing that the empirical
log-odds matrix gives the best explanation in a maximum-likelihood sense,
\emph{in the absence of any further structural restrictions}.

Hence, any log-odds matrix additionally restricted by structural assumptions will achieve a lower likelihood on the observed data.
However, in practice
the empirical log-odds matrix often has very poor predictive performance
because the estimate tends to have very large variance whose asymptotic is governed by the
number of times that entry is observed (which is practice is usually very small or even zero).

This variance may be reduced by regularising the complexity
of the estimated log-odds matrix. Common complexity measures of a
matrix are its matrix norms \citet{srebro2005rank}.
A natural choice is the nuclear norm or trace norm, which is a continuous surrogate for
rank and has found a wide range of machine-learning applications including matrix completion
\citep{candes2009exact,srebro2004maximum,pong2010trace}.

Recall, the trace norm of an $(n\times n)$ matrix $A$ is defined as
\[
\|A\|_{*}=\sum_{k=1}^{n}\sigma_{k}
\]
where $\sigma_{k}$ is the $k^{th}$ singular value of the matrix $A$.
The close relation to the rank of $A$ stems from the fact that the rank is the number of non-zero singular values.
When used in optimization, the trace norm behaves similar to the one-norm in LASSO type models,
yielding convex loss functions and forcing some singular values to be zero.

This principle can be used to obtain the following optimization program for regularized log-odds matrix estimation:
\begin{align*}
\min_{\bL} &\;\; \| \hat{\bL} - \bL \|_{F}^{2} + \lambda\|\bL\|_{*} \\
\mbox{s.t.}&\quad \bL+\bL^{\top}=0
\end{align*}
The first term is a Frobenius norm ``error term'', equivalent to a squared loss
$$\|\hat{\bL}-\bL\|_{F}^{2} = \sum_{i,j}(L_{ij}-\hat{L}_{ij})^{2},$$
instead of the log-likelihood function in order to ensure convexity of
the objective function.

There is a well-known bound on the trace of a matrix~\citep{srebro2004learning}:
For any $X\in\mathbb{R}^{n\times m}$, and $t\in\mathbb{R}$, $||X||_{*}\leq t$
if and only if there exists $A\in\mathbb{S}^{n}$ and $B\in\mathbb{S}^{m}$
such that $\left[\begin{array}{cc}
A & X\\
X^{\top} & B
\end{array}\right]\succeq0$ and $\frac{1}{2}\left(\Tr(A)+\Tr(B)\right)<t$. Using this bound,
we can introduce two auxiliary matrices $A$ and $B$ and solve an
equivalent problem:

\begin{align*}
\min_{A,B,\bL} &\;\; \|\hat{\bL}-\bL\|_{F}^{2}+\frac{\lambda}{2}\left(\Tr(A)+\Tr(B)\right) \\
\mbox{s.t.}&\quad
\left[\begin{array}{cc}
A & \bL\\
\bL^{\top} & B
\end{array}\right]\succeq0
\\
\mbox{and}&\quad \bL+\bL^{\top}=0
\end{align*}

This is a Quadratic Program with a positive semi-definite constraint
and a linear equality constraint. It can be efficiently solved by
the interior point method~\citet{vandenberghe1996semidefinite}, and
alternative algorithms for large scale settings also exist~\citep{mishra2013low}.

The estimation procedure can be generalized to model ternary match
outcomes. Without any structural assumption, the maximum likelihood
estimate for $p_{ij}[k]:=\text{P}(Y_{ij}=\text{k})$ is given by
\[
\hat{p}_{ij}[k]\coloneqq\frac{W_{ij}[k]}{N_{ij}}
\]
where $Y_{ij}$ is the ternary match outcome between team $i$ and
team $j$, and $k$ takes values in a discrete set of ordered levels.
$W_{ij}[k]$ is the number of matches between $i$ and $j$ in which
the outcome is $k$. $N_{ij}$ is the total number of matches between
the two teams as before.

We now define
$$
\bL_{ij}^{(1)}\coloneqq\log\left(\frac{p_{ij}[\mbox{win}]}{p_{ij}[\mbox{draw}] + p_{ij}[\mbox{lose}]}\right)\;\mbox{and}\;
\bL_{ij}^{(2)}\coloneqq\log\left(\frac{p_{ij}[\mbox{win}]+ p_{ij}[\mbox{draw}] }{p_{ij}[\mbox{lose}]}\right)
$$
The maximum likelihood estimate for $\bL_{ij}^{(1)}$ and $\bL_{ij}^{(2)}$
can be obtained by replacing $p_{ij}[k]$ with the
corresponding $\hat{p}_{ij}[k]$ in $\bL_{ij}^{(1)}$,
yielding maximum likelihood estimates $\hat{L}_{ij}^{(1)}$ and $\hat{L}_{ij}^{(2)}$.
As in Section~\ref{sub:Modeling-tenary-outcomes}, we make an implicit assumption of proportional odds
for which we will regularize, namely that $\bL_{ij}^{(2)}=\bL_{ij}^{(1)}+\phi$. For this, we obtain a new
convex objective function
\[
\min_{\bL,\phi} \|\hat{\bL}^{(1)}-\bL\|_{F}^{2}+\|\hat{\bL}^{(2)}-\bL-\phi\cdot \OOne\cdot \OOne^\top||_{F}^{2}+\lambda \|\bL\|_{*}.
\]
The optimal value of $\bL$ is a regularized estimate of $\bL_{ij}^{(1)}$, and $\bL + \phi\cdot \OOne\cdot \OOne^\top$
is a regularized estimate of $\bL_{ij}^{(2)}$.

The regularized log-odds matrix estimation method is quite experimental
as we have not established a mathematical proof for the error bound. Further research is also needed
to find an on-line update formula for this method.

We leave these as open questions for future investigations.

\section{Experiments\label{sec:Experiments}}

We perform two sets of experiments to validate the practical usefulness of
the novel structured log-odds models, including the Bradley-Terry-\'{E}l\H{o} model.

More precisely, we validate
\begin{enumerate}
\item[(i)] in the synthetic experiments in Section~\ref{sub:Synthetic-data} that the (feature-free) higher-rank models in Section~\ref{sec:specialcases} outperform the standard Bradley-Terry-\'{E}l\H{o} model
            if the generative process is higher-rank.
\item[(ii)] in real world experiments on historical English Premier League pairings, in Section~\ref{sub:Real-data-set},
structured log-odds models that use features as proposed in
Section~\ref{sub:covariate}, and the two-stage training method as proposed in Section~\ref{sub:Training-structured-log-odds} outperform methods that do not.
\end{enumerate}
In either setting, the methods outperform naive baselines, and their performance is similar to predictions derived from betting odds.

\subsection{Synthetic experiments\label{sub:Synthetic-data}}

In this section, we present the experiment results over synthetic
data sets. The goal of these experiments is to show that the newly
proposed structured log-odds models perform better than the original
\'{E}l\H{o} model when the data were generated following the new
models' assumptions. The experiments also show the validity of the
parameter estimation procedure.

The synthetic data are generated according to the assumptions of the
structured log-odds models (\ref{eq:model_summary}). To recap, the
data generation procedure is the following.
\begin{enumerate}
\item The binary match outcome $y_{ij}$ is sampled from a Bernoulli distribution
with success probability $p_{ij}$,
\item The corresponding log-odds matrix $L$ has a certain structure,
\item The match outcomes are sampled independently (there is no temporal
effect)\label{enu:The-match-outcomes}
\end{enumerate}
As the first step in the procedure, we randomly generate a ground
truth log-odds matrix with a certain structure. The structure depends
on the model in question and the matrix generation procedure is different
for different experiments. The match outcomes $y_{ij}$'s are sampled
independently from the corresponding Bernoulli random variables with
success probabilities $p_{ij}$ derived from the true log-odds matrix.

For a given ground truth matrix, we generate a validation set and
an independent test set in order to tune the hyper-parameter. The
hyper-parameters are the \textit{K factor} for the structured log-odds
models, and the \textit{regularizing strength $\lambda$} for regularized
log-odds matrix estimation. We perform a grid search to tune the hyper-parameter.
We choose the hyper-parameter to be the one that achieves the best
log-likelihood on the validation set. The model with the selected
hyper-parameter is then evaluated on the test set. This validation
setting is sound because of the independence assumption (\ref{enu:The-match-outcomes}).

The tuned model gives a probabilistic prediction for each match in
the test set. Based on these predictions, we can calculate the mean
log-likelihood or the mean accuracy on the test set. If two models
are evaluated on the same test set, the evaluation metrics for the
two models form a paired sample. This is because the metrics depend
on the specific test set.

In each experiment, we replicate the above procedure for many times.
In each replication, a new ground truth log-odds matrix is generated,
and the models are tuned and evaluated. Each replication hence produces
a paired sample of evaluation metrics because the metrics for different
models are conditional independent in the same replication.

We would like to know which model performs better given the data generation
procedure. This question can be answered by performing hypothesis
testing on paired evaluation metrics produced by the replications.
We will use the paired Wilcoxon test because of the violation of normality
assumption.

The experiments do not aim at comparing different training methods
(\ref{sub:Training-structured-log-odds}). Hence, all models in an
experiment are trained using the same method to enable an apple-to-apple
comparison. In experiments \ref{sub:fac2_exp} and \ref{sub:Rank-four-exp},
the structured log-odds models and the Bradley-Terry-\'{E}l\H{o} model are trained
by the online update algorithm. Experiment (\ref{sub:Regularized-log-odds-matrix-exp})
concerns about the regularized log-odds matrix estimation, whose online
update algorithm is yet to be derived. Therefore, all models in section
\ref{sub:Regularized-log-odds-matrix-exp} are trained using batch
training method.

The experiments all involve 47 teams \footnote{Forty-seven teams played in the English Premier league between 1993
and 2015}. Both validation and test set include four matches between each pair
of teams.

\subsubsection{Two-factor Bradley-Terry-\'{E}l\H{o} model\label{sub:fac2_exp}}

This experiment is designed to show that the two-factor
model is superior to the Bradley-Terry-\'{E}l\H{o} model if the true log-odds matrix
is a general rank-two matrix.

Components in the two factors $u$ and $v$ are independently generated
from a Gaussian distribution with $\mu=1$ and $\sigma=0.7$. The
true log-odds matrix is calculated as in equation \ref{eq:fac2_log_odds}
using the generated factors. The rest of the procedure is carried
out as described in section \ref{sub:Synthetic-data}. This procedure
is repeated for two hundred times.

\begin{figure}[H]
\begin{centering}
\includegraphics[scale=0.4]{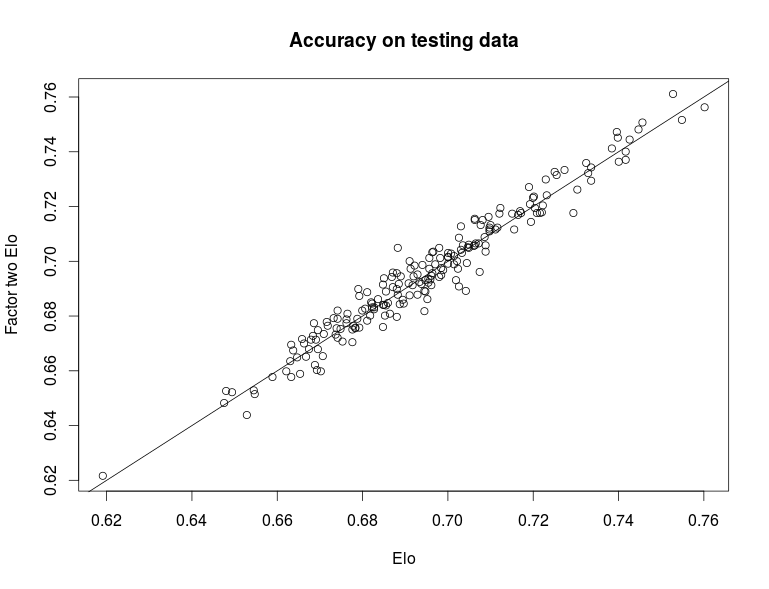}
\par\end{centering}

\caption{Each dot represents the testing accuracy in an experiment. The X-asis
shows the accuracy achieved by the \'{E}l\H{o} model while the Y-axis
shows the accuracy achieved by the two-factor \'{E}l\H{o}.\label{fig:Acc_fac2}}
\end{figure}

\begin{figure}[H]
\begin{centering}
\includegraphics[scale=0.4]{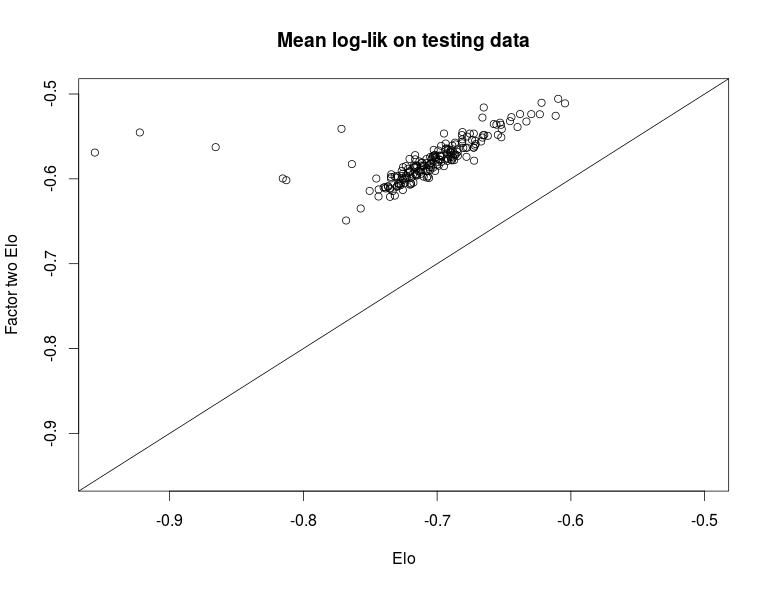}
\par\end{centering}

\caption{Each dot represents the mean log-likelihood on testing data in an
experiment. The X-asis shows the mean log-likelihood achieved by the
\'{E}l\H{o} model while the Y-axis shows the mean log-likelihood
achieved by the two-factor \'{E}l\H{o}.\label{fig:log_lik_fac2}}
\end{figure}

The two hundred samples of paired mean accuracy and paired mean log-likelihood
are visualized in figure \ref{fig:Acc_fac2} and \ref{fig:log_lik_fac2}.
Each point represents an independent paired sample.

Our hypothesis is that if the true log-odds matrix is a general rank-two
matrix, the two-factor \'{E}l\H{o} model is likely to perform better
than the original \'{E}l\H{o} model. We perform Wilcoxon test on
the paired samples obtained in the experiments. The two-factor \'{E}l\H{o}
model produces significantly better results in both metrics (one-sided
p-value is 0.046 for accuracy and less than $2^{-16}$ for mean log-likelihood).

\subsubsection{Rank-four Bradley-Terry-\'{E}l\H{o} model\label{sub:Rank-four-exp}}

These two experiments are designed to compare the rank-four \'{E}l\H{o}
model to the two-factor \'{E}l\H{o} model when the true log-odds
matrix is a rank-four matrix.

The first experiment considers the scenario when all singular values
of the true log-odds matrix are big. In this case, the best rank-two
approximation to the true log-odds matrix will give a relatively large
error because the third and fourth singular components cannot be recovered.
The log-odds matrices considered in this experiment takes the following
form
\begin{equation}
L=s_{1}\cdot u\cdot v^{\top}+s_{2}\cdot\theta\cdot\underline{1}^{\top}-s_{1}\cdot v\cdot u^{\top}-s_{2}\cdot\underline{1}\cdot\theta^{\top}\label{eq:rank4_exp}
\end{equation}
, where $s_{1}$ and $s_{2}$ are the two distinct singular values
and $\underline{1}$ is parallel to the vector of ones, and vector
$\underline{1}$ , $u$, $v$ and $\theta$ are orthonormal. This
formulation is based on the decomposition of a real antisymmetric
matrix stated in section \ref{sub:Motivation-and-definition}. The
true log-odds matrix $L$ has four non-zero singular values $s_{1}$,
$-s_{1}$, $s_{2}$ and $-s_{2}$. In the experiment, $s_{1}=25$
and $s_{2}=24$.

The rest of the data generation and validation setting is the same
as the experiments in section \ref{fig:log_lik_fac2}. The procedure
is repeated for 100 times. We applied the paired Wilcoxon test to
the 100 paired evaluation results. The test results support the hypothesis
that the rank-four \'{E}l\H{o} model performs significantly better
in both metrics (one-sided p-value is less than $2^{-16}$ for both
accuracy and mean log-likelihood).

In the second experiment, the components in factors $u$, $v$ and
$\theta$ are independently generated from a Gaussian distribution
with $\mu=1$ and $\sigma=0.7$. The log-odds matrix is then calculated
using equation \ref{eq:rank4_log_odds} directly. The factors are
no longer orthogonal and the second pair of singular values are often
much smaller than the first pair. In this case, the best rank-two
approximation will be close to the true log-odds matrix.

The procedure is repeated for 100 times again using the same data
generation and validation setting. Paired Wilcoxon test shows rank-four
\'{E}l\H{o} model achieves significantly higher accuracy on the test
data (one-sided p-value is 0.015), but the mean log-likelihood is
not significantly different (p-value is 0.81).

The results of the above two experiments suggest that the rank-four
\'{E}l\H{o} model will have significantly better performance when
the true log-odds matrix has rank four and it cannot be approximated
well by a rank-two matrix.

\subsubsection{Regularized log-odds matrix estimation\label{sub:Regularized-log-odds-matrix-exp}}

In the following two experiments, we want to compare the regularized
log-odds matrix estimation method with various structured log-odds
models.

To carry out regularized log-odds matrix estimation, we need to first
get an empirical estimate of log-odds on the training set. Since there
are only four matches between any pair of teams in the training data,
the estimate of log-odds often turn out to be infinity due to division
by zero. Therefore, I introduced a small regularization term in the
estimation of empirical winning probability $\hat{p}=\frac{n_{win}+\epsilon}{n_{total}+2\epsilon}$,
where $\epsilon$ is set to be 0.01. Then, we obtain the smoothed
log-odds matrix by solving the optimization problem described in section
\ref{sub:Regularized-log-odds-matrix}. A sequence of $\lambda$'s
are fitted, and the best one is chosen according to the log-likelihood
on the evaluation set. The selected model is then evaluated on the
testing data set.

Structured log-odds models with different structural assumptions are
used for comparison. We consider the \'{E}l\H{o} model, two-factor
\'{E}l\H{o} model, and rank-four \'{E}l\H{o} model. For each of
the three models, we first tune the hyper-parameter on a further split
of training data. Then, we evaluate the models with the best hyper-parameter
on the evaluation set and select the best model. Finally, we test
the selected model on the test set to produce evaluation metrics.
This experiment setting imitates the real application where we need
to select the model with best structural assumption.

In order to compare fairly with the trace norm regularization method
(which is currently a batch method), the structured log-odds models
are trained with batch method and the selected model is not updated
during testing.

In the first experiment, it is assumed that the structure of log-odds
matrix follows the assumption of the rank-four \'{E}l\H{o} model.
The log-odds matrix is generated using equation (\ref{eq:rank4_exp})
with $s_{1}=25$ and $s_{2}=2.5$. The data generation and hypothesis
testing procedure remains the same as previous experiments. Paired
Wilcoxon test is performed to examine the hypothesis that regularized
log-odds model produces higher out-of-sample log-likelihood. The testing
result is in favour of this hypothesis (p-value is less than $10^{-10}$).

In the second experiment, it is assumed that the structure of log-odds
matrix follows the assumption of the \'{E}l\H{o} model (section \ref{sub:The-Elo-model}).
The true \'{E}l\H{o} ratings are generated using a normal distribution
with mean $0$ and standard deviation $0.8$. Paired Wilcoxon test
shows that the out-of-sample likelihood is somewhat different between
the tuned regularized log-odds model and trace norm regularization
(two sided p-value = $0.09$).

The experiments show that regularized log-odds estimation can adapt
to different structures of the log-odds matrix by varying the regularization
parameter. The performance on simulated data set is not worse than
the tuned regularized log-odds model.

\subsection{Predictions on the English Premier League\label{sub:Real-data-set}}

\subsubsection{Description of the data set}

The whole data set under investigation consists of English Premier
League football matches from 1993-94 to 2014-15 season. There are
8524 matches in total. The data set contains the date of the match, the home team, the away
team, and the final scores for both teams. The English Premier League is chosen as a representative
as competitive team sports because of its high popularity. In each
season, twenty teams will compete against each other using the double
round-robin system: each team plays the others twice, once at the
home field and once as guest team. The winner of each match scores
three championship points. If the match draws, both teams score one
point. The final ranking of the teams are determined by the championship
points scored in the season. The team with the highest rank will be
the champion and the three teams with the lowest rank will move to
Division One (a lower-division football league) next season. Similarly,
three best performing teams will be promoted from Division One into
the Premier League each year. In the data set, 47 teams has played
in the Premier League. The data set is retrieved
from http://www.football-data.co.uk/.

  The algorithms are allowed to use all available information prior
to the match to predict the outcome of the match (win, lose, draw).

\subsubsection{Validation setting\label{sub:Tunning-and-validation}}

In the study of the real data set, we need a proper way to quantify
the predictive performance of a model. This is important for two reasons.
Firstly, we need to tune the hyper-parameters in the model by performing
model validation. The hyper-parameters that bring best performance
will be chosen. More importantly, we wish to compare the performance
of different types of models scientifically. Such comparison is impossible
without a quantitative measure on model performance.

It is a well-known fact that the errors made on the training data
will underestimate the model's true generalization error. The common
approaches to assess the goodness of a model include cross validation
and bootstrapping \citep{stone1974cross,efron1997improvements}. However,
both methods assume that the data records are statistically independent.
In particular, the records should not contain temporal structure.
In the literature, the validation for data with temporal structure
is largely an unexplored area. However, the independence assumption
is plausibly violated in this study and it is highly likely to affect
the result. Hence, we designed an set of ad-hoc validation methods
tailored for the current application.

The validation method takes two disjoint data sets, the training data
and the testing data. We concatenate the training and testing data
into a single data set and partition it into batches $\calD$
following the definitions given in \ref{sub:Training-structured-log-odds.online}. We then run Algorithm \ref{alg:batch_epoch_training} on $\calD$,
but only collect the predictions of matches in the testing data. Those
predictions are then compared with the real outcomes in the testing
data and various evaluation metrics can be computed.

The exact way to obtain batches $\calD$ will depend on the
training method we are using. In the experiments, we are mostly interested
in the repeated batch re-training method (henceforth batch training
method), the on-line training method and the two-stage training method.
For these three methods, the batches are defined as follows.
\begin{enumerate}
\item Batch training method: the whole training data forms the initial batch
$\calD_{0}$; the testing data is partitioned into similar-sized
batches based on time of the match.
\item On-line training method: all matches are partitioned into similar-sized
batches based on time of the match.
\item Two-stage method: the same as batch training method with a different
batch size on testing data.
\end{enumerate}

In general, a good validation setting should resemble the usage of
the model in practice. Our validation setting guarantees that no future
information will be used in making current predictions. It is also
naturally related to the training algorithm presented in \ref{sub:Training-structured-log-odds.online}.

\subsubsection{Prediction Strategy\label{sub:Prediction-Strategy}}

Most models in this comparative study have tunable hyper-parameters.
Those hyper-parameters are tuned using the above validation settings.
We split the whole data set into three disjoint subsets, the training set, the tuning set and the testing set.
The first match in the training set is the one between Arsenal and Coventry on 1993-08-04, and the first match in the tunning set is the one between Aston Villa and Blackburn on 2005-01-01. The first match in the testing data is the match between Stoke and Fulham on 2010-01-05, and the last match in the testing set is between Stoke and Liverpool on 2015-05-24. The testing set has 2048 matches in total.

In the tuning step, we supply the training set and the tuning set to the validation procedure as the training and testing data.
To find the best hyper-parameter, we perform a gird search and the hyper-parameter which
achieves the highest out-of-sample likelihood is chosen.
In theory, the batch size and epoch size are tunable hyper-parameters, but in the experiments we choose these parameters based on our prior knowledge. For the on-line and two-stage method, each individual match in testing data is regarded as a batch. The epoch size is chosen to be one. This reflects the usual update rule of the conventional \'{E}l\H{o} ratings: the ratings are updated immediately after the match outcome becomes available. For the batch training method, matches take place in the same quarter of the year are allocated to the same batch.

The model with the selected hyper-parameters is tested using
the same validation settings. The training data now consists of both training set and tuning set. The testing data is supplied with the testing set.

This prediction strategy ensures that the training-evaluating-testing
split is the same for all training methods, which means that the model
will be accessible to the same data set regardless of what training
method is being used. This ensures that we can compare different training methods fairly.

All the models will also be compared with a set of benchmarks. The first benchmark is a naive baseline which always predicts home team to win the match. The second benchmark is
constructed from the betting odds given by bookmakers. For each match, the bookmakers provide three odds for the three outcomes, win, draw and lose. The betting odds and the probability has the following relationship: $\text{P}=\frac{1}{\text{odds}}$. The probabilities implied by betting odds are used as prediction. However,
the bookmaker's odds will include a vigorish so the implied ``probability'' does
not sum to one. They are normalized by dividing each term with the sum to give the valid probability.
The historical odds are also obtained from http://www.football-data.co.uk/.

\subsubsection{Quantitative comparison for the evaluation metrics}

We use log-likelihood and accuracy on the testing data set as evaluation
metrics. We apply statistical hypothesis testing on the validation
results to compare the models quantitatively.

We calculate the log-likelihood on each test case for each model.
If we are comparing two models, the evaluation metrics for each test
case will form a paired sample. This is because test cases might be
correlated with each other and model's performance is independent
given the test case. The paired t-test is used to test whether there
is a significant difference in the mean of log-likelihood. We draw
independent bootstrap samples with replacement from the log-likelihood
values on test cases, and calculate the mean for each sample. We then
calculate the 95\% confidence interval for the mean log-likelihood
based on the empirical quantiles of bootstrapped means \citep{davison1997bootstrap}.
Five thousand bootstrap samples are used to calculate these intervals.

The confidence interval for accuracy is constructed assuming the model's
prediction for each test case, independently, has a probability $p$
to be correct. The reported 95\% confidence interval for Binomial
random variable is calculated from a procedure first given in \citet{clopper1934use}.
The procedure guarantees that the confidence level is at least 95\%,
but it may not produce the shortest-length interval.

\subsubsection{Performance of the structured log-odds model\label{sub:Performance-elo}}

We performed the tunning and validation of the structured log-odds
models using the method described in section \ref{sub:Tunning-and-validation}.
The following list shows all models examined by this experiment:
\begin{enumerate}
\item The Bradley-Terry-\'{E}l\H{o} model (section \ref{sub:The-Elo-model})
\item Two-factor Bradley-Terry-\'{E}l\H{o} model (section \ref{sub:The-structured-log-odds})
\item Rank-four Bradley-Terry-\'{E}l\H{o} model (section \ref{sub:The-structured-log-odds})
\item The Bradley-Terry-\'{E}l\H{o} model with score difference (section \ref{sub:Using-score-difference})
\item The Bradley-Terry-\'{E}l\H{o} model with two additional features (section \ref{sub:covariate})
\end{enumerate}
All models include a free parameter for home advantage (see section
\ref{sub:covariate}), and they are also able to capture the probability
of a draw (section \ref{sub:The-structured-log-odds}). We have introduced
two covariates in the fifth model. These two covariates indicate whether
the home team or away team is just promoted from Division One this
season. We have also tested the trace norm regularized log-odds model,
but as indicated in section \ref{sub:Regularized-log-odds-matrix}
the model still has many limitations for the application to the real
data. The validation results are summarized in table \ref{tab:elo-batch}
and table \ref{tab:elo-batch-1}.

The testing results help us understand the following two scientific
questions:
\begin{enumerate}
\item Which training method brings the best performance to structured log-odds
models?
\item Which type of structured log-odds model achieves best performance
on the data set?
\end{enumerate}
In order to answer the first question, we test the following hypothesis:
\begin{description}
\item [{(H1):}] Null hypothesis: for a certain model, two-stage training
method and online training method produce the same mean out-of-sample
log-likelihood. Alternative hypothesis: for a certain model two-stage
training method produces a higher mean out-of-sample log-likelihood
than online training method.
\end{description}
Here we compare the traditional on-line updating rule with the newly developed two-stage method.
The paired t-test is used to assess the above hypotheses. The p-values
are shown in table \ref{tab:test1}. The cell associated with the
\'{E}l\H{o} model with covariates are empty because the online training
method does not update the coefficients for features. The first columns
of the table gives strong evidence that the two-stage training method
should be preferred over online training. All tests are highly significant
even if we take into account the issue of multiple testing.

In order to answer the second question, we compare the four new models
with the Bradley-Terry-\'{E}l\H{o} model. The hypothesis is formulated as
\begin{description}
\item [{(H2):}] Null hypothesis: using the best training method, the new
model and the \'{E}l\H{o} model produce the same mean out-of-sample
log-likelihood. Alternative hypothesis: using the best training method,
the new model produces a higher mean out-of-sample log-likelihood
than the \'{E}l\H{o} model.
\end{description}
The p-values are listed in the last column of table \ref{tab:test1}.
The result also shows that adding more factors in the model does not
significantly improve the performance. Neither two-factor model nor
rank-four model outperforms the original Bradley-Terry-\'{E}l\H{o} model on the
testing data set. This might provide evidence and justification of
using the Bradley-Terry-\'{E}l\H{o} model on real data set. The model that uses
the score difference performs slightly better than the original Bradley-Terry-\'{E}l\H{o}
model. However, the difference in out-of-sample log-likelihood is
not statistically significant (the p-value for one-sided test is 0.24
for likelihood). Adding additional covariates about team promotion
significantly improves the Bradley-Terry-\'{E}l\H{o} model.

\begin{table}[H]
\begin{centering}
\begin{tabular}{|c|c|c|}
\hline
Type & H1 & H2\tabularnewline
\hline
\hline
\'{E}l\H{o} model & $7.8\times10^{-5}$ & -\tabularnewline
\hline
Two-factor model & $4.4\times10^{-14}$ & \textasciitilde{}1\tabularnewline
\hline
Rank-four model & $9.8\times10^{-9}$ & \textasciitilde{}1\tabularnewline
\hline
Score difference & $2.2\times10^{-16}$ & 0.235\tabularnewline
\hline
\'{E}l\H{o} model with covariates & - & 0.002\tabularnewline
\cline{2-3}
\end{tabular}
\par\end{centering}

\caption{Hypothesis testing on the structured log-odds model. The column ``Type''
specifies the type of the model; the remaining two columns shows the
one-sided p-values for the associated hypothesis\label{tab:test1}}
\end{table}

\begin{table}[H]
\begin{centering}
\begin{tabular}{|c|c|c|c|c|}
\hline
Type & Method & Acc  & 2.5\% & 97.5\%\tabularnewline
\hline
\hline
\multirow{2}{*}{Benchmark } & Home team win  & 46.07\%  & 43.93\%  & 48.21\% \tabularnewline
\cline{2-5}
 & Bet365 odds & \textbf{54.13\% } & 51.96\%  & 56.28\% \tabularnewline
\hline
\multirow{3}{*}{\'{E}l\H{o} model} & Two-stage & 52.40\%  & 50.23\%  & 54.56\% \tabularnewline
\cline{2-5}
 & Online & 52.16\% & 50.00\% & 54.32\%\tabularnewline
\cline{2-5}
 & Batch & 50.58\% & 48.41\% & 52.74\%\tabularnewline
\hline
\multirow{3}{*}{Two-factor model} & Two-stage & 51.30\%  & 49.13\%  & 53.46\% \tabularnewline
\cline{2-5}
 & Online & 50.34\% & 48.17\% & 52.50\%\tabularnewline
\cline{2-5}
 & Batch & 50.86\% & 48.69\% & 53.03\%\tabularnewline
\hline
\multirow{3}{*}{Rank-four model} & Two-stage & 51.34\% & 49.17\%  & 53.51\% \tabularnewline
\cline{2-5}
 & Online & 50.34\% & 48.17\% & 52.50\%\tabularnewline
\cline{2-5}
 & Batch & 50.58\% & 48.41\% & 52.74\%\tabularnewline
\hline
\multirow{3}{*}{Score difference} & Two-stage & 52.59\% & 50.42\% & 54.75\%\tabularnewline
\cline{2-5}
 & Online & 47.17\% & 45.01\% & 49.34\%\tabularnewline
\cline{2-5}
 & Batch & 51.10\% & 48.93\% & 53.27\%\tabularnewline
\hline
\multirow{2}{*}{\'{E}l\H{o} model with covariates} & Two-stage & \textbf{52.78\%} & 50.61\% & 54.95\%\tabularnewline
\cline{2-5}
 & Batch & 50.86\% & 48.69\% & 53.03\%\tabularnewline
\hline
Trace norm regularized model & Batch & 45.89\% & 43.54\% & 48.21\%\tabularnewline
\hline
\end{tabular}
\par\end{centering}

\caption{Structured log-odds model's accuracy on testing data. The column ``Type''
specifies the type of the model; the column ``Method'' specifies
the training method. Testing accuracy is given in the column ``Acc''.
The last two columns gives the 95\% confidence interval for testing
accuracy \label{tab:elo-batch}}
\end{table}

\begin{table}[H]
\begin{centering}
\begin{tabular}{|c|c|c|c|c|}
\hline
Type & Method & Mean log-loss & 2.5\% & 97.5\%\tabularnewline
\hline
\hline
Benchmark & Bet365 odds & \textbf{-0.9669} & -0.9877 & -0.9460\tabularnewline
\hline
\multirow{3}{*}{\'{E}l\H{o} model} & Two-stage & -0.9854 & -1.0074 & -0.9625\tabularnewline
\cline{2-5}
 & Online & -1.0003 & -1.0254 & -0.9754\tabularnewline
\cline{2-5}
 & Batch & -1.0079 & -1.0314 & -0.9848\tabularnewline
\hline
\multirow{3}{*}{Two-factor model} & Two-stage & -1.0058 & -1.0286 & -0.9816\tabularnewline
\cline{2-5}
 & Online & -1.0870 & -1.1241 & -1.0504\tabularnewline
\cline{2-5}
 & Batch & -1.0158 & -1.0379 & -0.9919\tabularnewline
\hline
\multirow{3}{*}{Rank-four model} & Two-stage & -1.0295 & -1.0574 & -1.0016\tabularnewline
\cline{2-5}
 & Online & -1.1024 & -1.0638 & -1.1421\tabularnewline
\cline{2-5}
 & Batch & -1.0078 & -1.0291 & -0.9860\tabularnewline
\hline
\multirow{3}{*}{Score difference} & Two-stage & -0.9828 & -1.0034 & -0.9623\tabularnewline
\cline{2-5}
 & Online & -1.1217 & -1.1593 & -1.0833\tabularnewline
\cline{2-5}
 & Batch & -1.0009 & -1.0206 & -0.9802\tabularnewline
\hline
\multirow{2}{*}{\'{E}l\H{o} model with covariates} & Two-stage & \textbf{-0.9807} & -1.0016 & -0.9599\tabularnewline
\cline{2-5}
 & Batch & -1.0002 & -1.0204 & -0.9798\tabularnewline
\hline
\end{tabular}
\par\end{centering}

\caption{Structured log-odds model's mean log-likelihood on testing data. The
column ``Type'' specifies the type of the model; the column ``Method''
specifies the training method. Mean out-of-sample log-likelihood is
given in the column ``Mean log-loss''. The last two columns gives
the 95\% confidence interval for mean out-of-sample log-likelihood\label{tab:elo-batch-1}}
\end{table}

\subsubsection{Performance of the batch learning models}

This experiment compares the performance of batch learning models.
The following list shows all models examined by this experiment:
\begin{enumerate}
\item GLM with elastic net penalty using multinomial link function
\item GLM with elastic net penalty using ordinal link function
\item Random forest
\item Dixon-Coles model
\end{enumerate}
The first three models are machine learning models that can be trained
on different features. The following features are considered in this
experiment:
\begin{enumerate}
\item Team id: the identity of home team and away team
\item Ranking: the team's current ranking in Championship points and goals
\item VS: the percentage of time that home team beats away team in last
3, 6, and 9 matches between them
\item Moving average: the moving average of the following monthly features
using lag 3, 6, 12, and 24

\begin{enumerate}
\item percentage of winning at home
\item percentage of winning away
\item number of matches at home
\item number of matches away
\item championship points earned
\item number of goals won at home
\item number of goals won away
\item number of goals conceded at home
\item number of goals conceded away
\end{enumerate}
\end{enumerate}
The testing accuracy and out-of-sample log-likelihood are summarized
in table \ref{tab:acc_batch} and table \ref{tab:log-lik-batch}.
All models perform better than the baseline benchmark, but no model
seems to outperform the state-of-the-art benchmark (betting odds).

We applied statistical testing to understand the following questions
\begin{enumerate}
\item Does the GLM with ordinal link function perform better than the GLM
with multinomial link function?
\item Which set of features are most useful to make prediction?
\item Which model performs best among GLM, Random forest, and Dixon-Coles
model?
\end{enumerate}
For question one, we formulate the hypothesis as:
\begin{description}
\item [{(H3):}] Null hypothesis: for a given set of feature, the GLM with
ordinal link function and the GLM with multinomial link function produce
the same mean out-of-sample log-likelihood. Alternative hypothesis:
for a given set of feature, the mean out-of-sample log-likelihood
is different for the two models.
\end{description}
The p-values for these tests are summarized in table \ref{tab:p-values-for-H5}.
In three out of four scenarios, the test is not significant. There
does not seem to be enough evidence against the null hypothesis. Hence,
we retain our believe that the GLM with different link functions have
the same performance in terms of mean out-of-sample log-likelihood.

\begin{table}
\begin{centering}
\begin{tabular}{|c|c|}
\hline
Features  & p-value\tabularnewline
\hline
\hline
Team\_id only  & 0.148\tabularnewline
\hline
Team\_id and ranking  & 0.035\tabularnewline
\hline
Team\_id and VS & 0.118\tabularnewline
\hline
Team\_id and MA & 0.121\tabularnewline
\hline
\end{tabular}
\par\end{centering}

\caption{p-values for H3\label{tab:p-values-for-H5}}
\end{table}

For question two, we observe that models with the moving average feature
have achieved better performance than the same model trained with
other features. We formulate the hypothesis as:
\begin{description}
\item [{(H4):}] Null hypothesis: for a given model, the moving average
feature and an alternative feature set produce the same mean out-of-sample
log-likelihood. Alternative hypothesis: for a given model, the mean
out-of-sample log-likelihood is higher for the moving average feature.
\end{description}
The p-values are summarized in table \ref{tab:p-values-for-H6}. The
tests support our believe that the moving average feature set is the
most useful one among those examined in this experiment.

\begin{table}
\begin{centering}
\begin{tabular}{|c|c|c|}
\hline
Features  & GLM1 & GLM2\tabularnewline
\hline
\hline
Team\_id only  & $2.7\times10^{-12}$ & $5.3\times10^{-8}$\tabularnewline
\hline
Team\_id and ranking  & $1.2\times10^{-9}$ & $3.7\times10^{-6}$\tabularnewline
\hline
Team\_id and VS & 0.044 & 0.004\tabularnewline
\hline
\end{tabular}
\par\end{centering}

\caption{p-values for H4: the column ``Features'' are the alternative features
compared with the moving average features. The next two columns contain
the p-values for the GLM with multinomial link function (GLM1) and
the GLM with ordinal link function (GLM2) \label{tab:p-values-for-H6}}
\end{table}

Finally, we perform comparison among different models. The comparisons
are made between the GLM with multinomial link function, Random forest,
and Dixon-Coles model. The features used are the moving average feature
set. The p-values are summarized in table \ref{tab:p-values-for-H7}.
The tests detect a significant difference between GLM and Random forest,
but the other two pairs are not significantly different. We apply
the p-value adjustment using Holm's method in order to control family-wise
type-one error \citep{sinclair2013alpha}. The adjusted p-values are
not significant. Hence, we retain our belief that the three models
have the same predictive performance in terms of mean out-of-sample
log-likelihood.

\begin{table}
\begin{centering}
\begin{tabular}{|c|c|c|}
\hline
Comparison & p-value & adjusted\tabularnewline
\hline
\hline
GLM and RF & 0.03 & 0.08\tabularnewline
\hline
GLM and DC & 0.48 & 0.96\tabularnewline
\hline
DC and RF & 0.54 & 0.96\tabularnewline
\hline
\end{tabular}
\par\end{centering}

\caption{p-values for model comparison: the column ``Comparison'' specifies
which two models are being compared. ``RF'' stands for Random forest;
``DC'' stands for the Dixon-Coles model. The column ``p-value''
contains the two-sided p-value of the corresponding paired t-test.
The column ``adjusted'' shows the adjusted p-values for multiple
testing\label{tab:p-values-for-H7}}
\end{table}

\begin{table}
\begin{centering}
\begin{tabular}{|c|c|c|c|c|}
\hline
Models & Features  & Acc  & 2.5\%  & 97.5\% \tabularnewline
\hline
\hline
\multirow{2}{*}{Benchmark } & Home team win  & 46.07\%  & 43.93\%  & 48.21\% \tabularnewline
\cline{2-5}
 & Bet365 odds & 54.13\%  & 51.96\%  & 56.28\% \tabularnewline
\hline
\multirow{4}{*}{GLM1} & Team\_id only  & 50.05\%  & 47.88\%  & 52.22\% \tabularnewline
\cline{2-5}
 & Team\_id and ranking  & 50.62\%  & 48.45\%  & 52.79\% \tabularnewline
\cline{2-5}
 & Team\_id and VS & 51.25\%  & 49.08\%  & 53.41\% \tabularnewline
\cline{2-5}
 & Team\_id and MA & \textbf{52.69\% } & 50.52\%  & 54.85\% \tabularnewline
\hline
\multirow{4}{*}{GLM2} & Team\_id only  & 50.67\%  & 48.52\%  & 52.82\% \tabularnewline
\cline{2-5}
 & Team\_id and ranking  & 50.24\%  & 48.09\%  & 52.38\% \tabularnewline
\cline{2-5}
 & Team\_id and VS  & 51.92\%  & 49.75\%  & 54.08\% \tabularnewline
\cline{2-5}
 & Team\_id and MA  & 52.93\%  & 50.76\%  & 55.09\% \tabularnewline
\hline
RF & Team\_id and MA  & 52.06\% & 49.89\% & 54.23\%\tabularnewline
\hline
Dixon-Coles  & - & 52.54\%  & 50.40\%  & 54.68\% \tabularnewline
\hline
\end{tabular}
\par\end{centering}

\caption{Testing accuracy for batch learning models: The column ``Type''
specifies the type of the model; ``GLM1'' refers to the GLM with
multinomial link function, and ``GLM2'' refers to the GLM with ordinal
link function. column ``Models'' specifies the model, and the column
``Features'' specifies the features used to train the model. Testing
accuracy is given in the column ``Acc''. The last two columns gives
the 95\% confidence interval for testing accuracy.\label{tab:acc_batch}}
\end{table}

\begin{table}
\begin{centering}
\begin{tabular}{|c|c|c|c|c|}
\hline
Models & Features  & Mean log-loss  & 2.5\% & 97.5\% \tabularnewline
\hline
\hline
\multirow{1}{*}{Benchmark} & Bet365 odds & -0.9669 & -0.9877 & -0.9460\tabularnewline
\hline
\multirow{4}{*}{GLM1} & Team\_id only  & -1.0123 & -1.0296 & -0.9952\tabularnewline
\cline{2-5}
 & Team\_id and ranking  & -1.0006 & -1.0175 & -0.9829\tabularnewline
\cline{2-5}
 & Team\_id and VS & -0.9969 & -1.0225 & -0.9721\tabularnewline
\cline{2-5}
 & Team\_id and MA & \textbf{-0.9797} & -0.9993 & -0.9609\tabularnewline
\hline
\multirow{4}{*}{GLM2} & Team\_id only  & -1.0184 & -1.0399 & -0.9964\tabularnewline
\cline{2-5}
 & Team\_id and ranking  & -1.0097 & -1.0317 & -0.9874\tabularnewline
\cline{2-5}
 & Team\_id and VS  & -1.0077 & -1.0338 & -0.9813\tabularnewline
\cline{2-5}
 & Team\_id and MA  & -0.9838 & -1.0028 & -0.9656\tabularnewline
\hline
RF & Team\_id and MA  & -0.9885 & -1.0090 & -0.9683\tabularnewline
\hline
Dixon-Coles  & - & -0.9842 & -1.0076 & -0.9610\tabularnewline
\hline
\end{tabular}
\par\end{centering}

\caption{out-of-sample log-likelihood for batch learning models: The column
``Type'' specifies the type of the model; ``GLM1'' refers to the
GLM with multinomial link function, and ``GLM2'' refers to the GLM
with ordinal link function. the column ``Models'' specifies the
model, and the column ``Features'' specifies the features used to
train the model. Mean out-of-sample log-likelihood is given in the
column ``Mean log-loss''. The last two columns gives the 95\% confidence
interval for mean out-of-sample log-likelihood.\label{tab:log-lik-batch}}
\end{table}

\subsection{Fairness of the English Premier League ranking}

``Fairness'' as a concept is statistically undefined and due to its subjectivity is not empirical unless based on peoples' opinions.
The latter may wildly differ and are not systematically accessible from our data set or in general.

Hence we will base our study of the Premier League ranking scheme's ``fairness'' on a surrogate derived
from the following plausibility considerations:
Ranking in any sport should plausibly be based on the participants' skill in competing in official events of that sport.
By definition the outcomes of such events measure the skill in competing at the sport, distorted by a possible component of ``chance''.
The ranking, derived exclusively from such outcomes, will hence also be determined by the so-measured skills and a component of ``chance''.

A ranking system may plausibly be considered fair if the final ranking is only minimally affected by whatever constitutes ``chance'',
while accurately reflecting the ordering of participating parties in terms of skill, i.e., of being better at the game.

Note that such a definition of fairness is disputable, but it may agree with the general intuition when ranking players of games with a strong chance
component such as card or dice games, where cards dealt or numbers thrown in a particular game should, intuitively, not affect a player's rank,
as opposed to the player's skills of making the best out of a given dealt hand or a dice throw.

Together with the arguments from Section~\ref{sub:intro_one} which argue for predictability-in-principle surrogating skill,
and statistical noise surrogating chance, fairness may be surrogated as the stability of the ranking under the best possible prediction
that surrogates the ``true odds''.
In other words, if we let the same participants, under exactly the same conditions, repeat the whole season, and all that changes is
the dealt cards, the thrown numbers, and similar possibly unknown occurrences of ``chance'',
are we likely to end up with the same ranking as the first time?

While of course this experiment is unlikely to be carried out in real life for most sports, the best possible prediction which is surrogated by the prediction by the best accessible predictive model yields a statistically justifiable estimate for the outcome of such a hypothetical real life experiment.

To obtain this estimate, we consider the as the ``best accessible predictive model''
the Bradley-Terry-\'{E}l\H{o} model with features, learnt by the two-stage update rule (see Section~\ref{sub:Performance-elo}),
yielding a probabilistic prediction for every game in the season.
From these predictions, we may independently sample match outcomes and
final rank tables according to the official scoring and ranking rules.

\begin{figure}[H]
\begin{centering}
\includegraphics[scale=0.8]{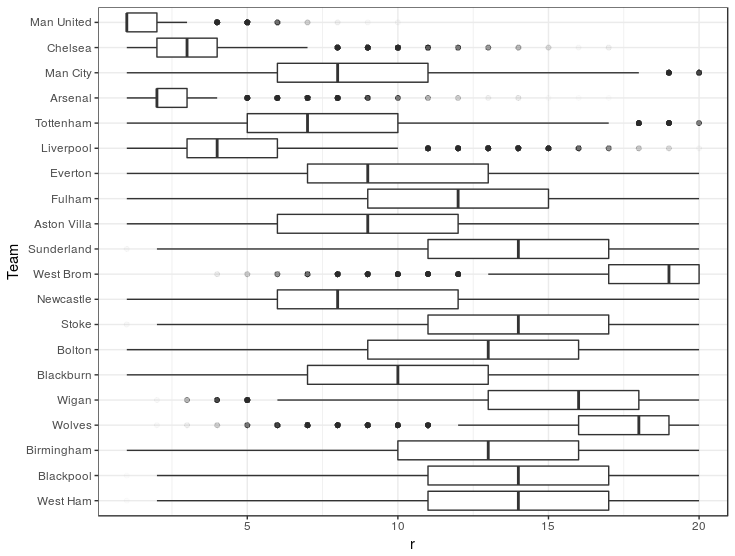}
\par\end{centering}

\caption{Estimated probability for each team participating in the English Premier League season 2010-2011
to obtain the given final rank. Rows are indexed by the different teams in the Premier League of 2010-2011, ordered descendingly by their actual final rank.
The x-axis is indexed by the possible ranks from 1 (best) to 20 (worst). The horizontal box-plots are obtained from a Monte-Carlo-sample
from 10.000 of the predictive ranking distribution; boxes depict estimates the 25\%, 50\% and 75\%
quantiles of the predictive distribution's Monte Carlo estimate, with whiskers being min/max or 1.5IQR.  \label{fig:final_ranking}}
\end{figure}

Figure~\ref{fig:final_ranking} shows estimates for the distribution or ranks of Premier League teams participating in the 2010 season.

It may be observed that none of the teams, except Manchester United, ends up with the same rank they achieved in reality in more than 50\% of the cases.
For most teams, the middle 50\% are spread over 5 or more ranks, and for all teams, over 2 or more.

From a qualitative viewpoint, the outcome for most teams appears very random, hence the allocation of the final rank seems qualitatively similar to a game of chance
notable exceptions being Manchester United and Chelsea whose true final rank is similar to a narrow expected/predicted range. It is also worthwhile noting that Arsenal
has been predicted/expected among the first three with high confidence, but eventually was ranked fourth.

The situation is qualitatively similar for later years, though not shown here.

\newpage

\section{Discussion and Summary\label{sec:Summary-and-Conclusion}}

We discuss our findings in the context of our questions regarding prediction
of competitive team sports and modelling of English Premier League outcomes, compare Section~\ref{sec:Questions}

\subsection{Methodological findings}

As the principal methodological contribution of this study, we have formulated the Bradley-Terry-\'{E}l\H{o} model
in a joint form, which we have extended to the flexible class of structured log-odds models.
We have found structured log-odds models to be potentially useful in the following way:

\begin{enumerate}
\item[(i)] The formulation of the Bradley-Terry-\'{E}l\H{o} model as a parametric model within a supervised on-line setting solves a number of open issues of the heuristic \'{E}l\H{o} model, including setting of the K-factor and new players/teams.
\item[(ii)] In synthetic experiments, higher rank \'{E}l\H{o} models outperform the Bradley-Terry-\'{E}l\H{o} model in predicting competitive outcomes if the generative truth is higher rank.
\item[(iii)] In real world experiments on the English Premier league, we have found that the extended capability of structured log-odds models to make use of features is useful as it allows better prediction of outcomes compared to not using features.
\item[(iv)] In real world experiments on the English Premier league, we have found that our proposed two-stage training strategy for on-line learning with structured log-odds models is useful as it allows better prediction of outcomes compared to using standard on-line strategies or batch training.
\end{enumerate}

We would like to acknowledge that many of the mentioned suggestions and extensions are already found in existing literature, while, similar to the Bradley-Terry and \'{E}l\H{o} models in which parsimonious parametric form and on-line learning rule have been separated, those ideas usually appear without being joint to a whole.
We also anticipate that the highlighted connections to generalized linear models, low-rank matrix completion and neural networks may prove fruitful in future investigations.

\subsection{Findings on the English Premier League}

The main empirical on the English Premier League data may be described as follows.

\begin{enumerate}
\item[(i)] The best predictions, among the methods we compared, are obtained from a structured log-odds model with rank one and added covariates (league promotion), trained via the two-stage strategy. Not using covariates or the batch training method makes the predictions (significantly) worse (in terms of out-of-sample likelihood).
\item[(ii)] All our models and those we adapted from literature were outperformed by the Bet365 betting odds.
\item[(iii)] However, all informed models were very close to each other and the Bet 365 betting odds in performance and not much better than the uninformed baseline of team-independent home team win/draw/lose distribution.
\item[(iv)] Ranking tables obtained from the best accessible predictive model (as a surrogate for the actual process by which it is obtained, i.e., the games proper) are, qualitatively, quite random, to the extent that most teams may end up in wildly different parts of the final table.
\end{enumerate}

While we were able to present a parsimonious and interpretable state-of-art model for outcome prediction for the English Premier League,
we found it surprising how little the state-of-art improves above an uninformed guess which already predicts almost half the (win/lose/draw) outcomes correctly,
while differences between the more sophisticated methods range in the percents.

Given this, it is probably not surprising that a plausible surrogate for humanity's ``secret'' or non-public knowledge of competitive sports prediction,
the Bet365 betting odds, is not much better either. Note that this surrogate property is strongly plausible from noticing that offering odds leading to a worse prediction
leads to an expected loss in money, hence the market indirectly forces bookmakers to disclose their best prediction\footnote{
The expected log-returns of a fractional portfolio where a fraction $q_i$ of the money is bet on outcome $i$ against a bookmaker whose odds correspond to probabilities $p_i$
are $\EE [L_\ell (p,Y)] - \EE[L_\ell (q,Y)] - c$ where $L_\ell$ is the log-loss and $c$ is a vigorish constant. In this utility quantifier, portfolio composition and bookmaker odds
are separated, hence in a game theoretic adversarial minimax/maximin sense, the optimal strategies consist in the bookmaker picking $p$ and the player picking $q$ to be their best possible/accessible prediction, where ``best'' is measured through expected log-loss (or an estimate thereof). Note that this argument does not take into account behavioural aspects
or other utility/risk quantifiers such as a possible risk premium, so one should consider it only as an approximation, though one that is plausibly sufficient for the qualitative discussion in-text.
}.
Thus, the continued existence of betting companies hence may lead to the belief that this is possibly rather due to
predictions of ordinary people engaged in betting that are worse than uninformed, rather than betting companies' capability of predicting better.
Though we have not extensively studied betting companies empirically, hence this latter belief is entirely conjectural.

Finally, the extent to which the English Premier League is unpredictable raises an important practical concern:
influential factors cannot be determined from the data if prediction is impossible, since by recourse to the scientific method
assuming an influential factor is one that improves prediction.
Our results above allow to definitely conclude only three such factors which are observable, namely a general ``good vs bad'' quantifier for whatever one may consider as a team's ``skills'', which of the teams is at home, and the fact whether the team is new to the league.
As an observation, this is not very deep or unexpected - the surprising aspect is that we were not able to find evidence for more.
On a similar note, it is surprising how volatile a team's position in the final ranking tables seems to be, given the best prediction we were able to achieve.

Hence it may be worthwhile to attempt to understand the possible sources of the observed nigh-unpredictability.
On one hand, it can simply be that the correct models are unknown to us and the right data to make a more accurate prediction have been disregarded by us.
Though this is made implausible by the observation that the betting odds are similarly bad in predicting, which is somewhat surprising as we have not used much of possibly available detail data such as in-match data and/or player data (which are heavily advertised by commercial data providers these days).
On the other hand, unpredictability may be simply due to a high influence of chance inherent to English Premier League games,
similar to a game of dice that is not predictable beyond the correct odds.
Such a situation may plausibly occur if the ``skill levels'' of all the participating teams are very close - in an extreme case,
where 20 copies of the same team play against each other, the outcome would be entirely up to chance as the skills match exactly, no matter how good or bad these are.
Rephrased differently, a game of skill played between two players of equal skill becomes a game of chance.
Other plausible causes of the situation is that the outcome a Premier League game is more governed by chance and coincidence than by skills in the first place,
or that there are unknown influential factors which are unobserved and possibly distinct from both chance or playing skills.
Of course, the mentioned causes do not exclude each other and may be present in varying degrees not determinable from the data considered in this study.

From a team's perspective, it may hence be interesting to empirically re-evaluate measures that are very costly or resource consuming under the aspect of
predictive influence in a similar analysis, say.

\subsection{Open questions}
A number of open research questions and possible further avenues of investigation have already been pointed out in-text.
We summarize what we believe to be the most interesting avenues for future research:

\begin{enumerate}
\item[(i)] A number of parallels have been highlighted between structured log-odds models and neural networks.
            It would be interesting to see whether adding layers or other ideas of neural network flavour are beneficial in any application.
\item[(ii)] The correspondence to low-rank matrix completion has motivated a nuclear norm regularized algorithm; yielding acceptable results in a synthetic scenario, the algorithm did not perform better than the baseline on the Premier League data. While this might be due to the above-mentioned issues with that data, general benefits of this alternative approach to structured log-odds models may be worth studying - as opposed to training approaches closer to logistic regression and neural networks.
\item[(iii)] The closeness to low-rank matrix completion also motivates to study identifiability and estimation variance bounds on particular entries of the log-odds matrix, especially in a setting where pairings are not independently or uniformly sampled.
\item[(iv)] While our approach to structured log-odds is inherently parametric, it is not fully Bayesian - though naturally, the benefit of such an approach may be interesting to study.
\item[(v)] We did not investigate in too much detail the use of features such as player data, and structural restrictions on the feature coefficient matrices and tensors. Doing this, not necessarily in the context of the English Premier League, might be worthwhile, though such a study would have to rely on good sources of added feature data to have any practical impact.
\end{enumerate}

On a more general note, the connection between neural networks and low-rank or matrix factorization principles apparent in this work may also be an interesting direction to explore,
not necessarily in a competitive outcome prediction context.

\bibliographystyle{plainnat}
\bibliography{elo}

\end{document}